\documentclass[journal]{IEEEtran}
\usepackage{amsmath,epsfig,amsfonts}
\usepackage{makecell}
\usepackage{epstopdf}
\usepackage{graphicx}
\usepackage{subfigure}

\ifCLASSINFOpdf
\else
\fi

\begin{document}
%
\title{Rank-Consistency Deep Hashing for Scalable Multi-Label Image Search}
%
%
%
\author{Cheng~Ma,
        Jiwen~Lu,~\IEEEmembership{Senior~Member,~IEEE,}
        and~Jie~Zhou,~\IEEEmembership{Senior~Member,~IEEE}
\thanks{\IEEEcompsocthanksitem 
This work was supported in part by the National Key Research and Development Program of China under Grant 2017YFA0700802, in part by the National Natural Science Foundation of China under Grant 61822603, Grant U1813218, Grant U1713214, and Grant 61672306, in part by Beijing Academy of Artificial Intelligence (BAAI) under Grant BAAI2020ZJ0202, in part by a grant from the Institute for Guo Qiang, Tsinghua University, in part by the Shenzhen Fundamental Research Fund (Subject Arrangement) under Grant JCYJ20170412170602564, and in part by Tsinghua University Initiative Scientific Research Program. \emph{(Corresponding author: Jiwen Lu)}.

Cheng Ma and Jiwen Lu are with the State Key Lab of Intelligent Technologies and Systems, Beijing National Research Center for Information Science and Technology (BNRist) and the Department of Automation, Tsinghua University, Beijing, 100084, China. Email: macheng17@mails.tsinghua.edu.cn; lujiwen@tsinghua.edu.cn.

Jie Zhou is with the State Key Lab of Intelligent Technologies and Systems, Beijing National Research Center for Information Science and Technology (BNRist), the Department of Automation, Tsinghua University, Beijing 100084, China, and also with the Tsinghua Shenzhen International Graduate School, Tsinghua University, Shenzhen 518055, China. Email: jzhou@tsinghua.edu.cn.}}

%



\maketitle

\begin{abstract}
As hashing becomes an increasingly appealing technique for large-scale image retrieval, multi-label hashing is also attracting more attention for the ability to exploit multi-level semantic contents. In this paper, we propose a novel deep hashing method for scalable multi-label image search. Unlike existing approaches with conventional objectives such as contrast and triplet losses, we employ a rank list, rather than pairs or triplets, to provide sufficient global supervision information for all the samples. Specifically, a new rank-consistency objective is applied to align the similarity orders from two spaces, the original space and the hamming space. 
A powerful loss function is designed to penalize the samples whose semantic similarity and hamming distance are mismatched in two spaces. Besides, a multi-label softmax cross-entropy loss is presented to enhance the discriminative power with a concise formulation of the derivative function. In order to manipulate the neighborhood structure of the samples with different labels, we design a multi-label clustering loss to cluster the hashing vectors of the samples with the same labels by reducing the distances between the samples and their multiple corresponding class centers.  The state-of-the-art experimental results achieved on three public multi-label datasets, MIRFLICKR-25K, IAPRTC12 and NUS-WIDE, demonstrate the effectiveness of the proposed method. 
\end{abstract}


%
\IEEEpeerreviewmaketitle

\section{Introduction}
%
%
%
%
\IEEEPARstart{L}{arge-scale} image retrieval has been very important in the field of computer vision and multimedia computing with the arrival of the era of big data. Consequently, worldwide researchers pay more and more attention to efficient image representation methods, such as LBP~\cite{ahonen2006face,ojala2002multiresolution,guo2010completed}, GIST~\cite{oliva2001modeling} and tree-based techniques~\cite{friedman2018algorithm,guttman1984r}. Besides, hashing methods~\cite{lin2015supervised,liu2015multi,wang2010semi,chen2016nonlinear,li2015feature,chen2017nonlinear,nguyen2014supervised,liong2018cross,yuan2018reconstruction,liong2017learning,chen2018order} have been popular to solve this problem on account of the storage and indexing efficiency. They aim to learn several hashing functions to extract compact binary codes in low-dimension space from raw images in high-dimension space. Then through the learned binary codes in the hamming space, the image similarity search can be achieved efficiently. Hence it is important that semantically similar images have similar binary codes with small hamming distances.

A number of hashing methods have been proposed. These methods can be divided into two categories, data-independent methods including Locality Sensitive Hashing (LSH)~\cite{gionis1999similarity} and data-dependent methods such as Spectral Hashing~\cite{weiss2009spectral} and Iterative Quantization (ITQ)~\cite{gong2013iterative}. However, many early methods are inadequate to capture multi-hierarchical semantic information since they are shallow and only consider linear projection. Meanwhile, they are usually under some assumptions which contribute to the obtainment of the solutions, but are hard to be satisfied in practice. Hence various deep hashing methods~\cite{srivastava2012multimodal,liong2017cross,zhuang2016fast,zhang2015bit,erin2015deep,lu2017deep} have been proposed based on the development of deep convolutional neural networks (CNNs). The breakthrough on representation learning promotes the capacity for scalable image retrieval by hashing.  

Moreover, an image is often associated with more than one concept in the real world, so that single-label image retrieval methods are not capable enough of meeting the demand of flexibility and accuracy. Hence more attention is drawn to the task of multi-label image retrieval due to its practicability. It is more difficult on account of the discrepancy in attentional regions and the variability of the components and tag numbers in one image~\cite{ji2014query,li2013learning,song2015top}. Different from single-label hashing, multi-label hashing has to concentrate on multi-level semantic information contained in images associated with multiple labels. Therefore, multi-label hashing may have the ability to capture both holistic and local knowledge of images, which is more and more important for large-scale image retrieval in the future. 

However, there are two major constraints with existing multi-label hashing methods. (1) Most multi-label hashing methods are based on pairwise similarity~\cite{zhang2018instance,wu2017deep} and triplet loss~\cite{zhao2015deep,lai2016instance}. Such methods suffer from data expansion when the pairs and triplets are sampled from the training dataset. Meanwhile, the multi-label supervision signals may lead to local optimum due to the limited information provided by the images in pairs or triplets. (2) The classification task has been involved in single-label hashing methods~\cite{yao2016deep,li2017deep} to improve the semantically discriminative ability of the hashing codes. However, few multi-label hashing methods have appropriately made use of the discrimination and polymerization power of the learned features, which leads to the bottleneck of the progress in multi-label image retrieval.

In this paper, we propose a rank-consistency deep hashing method (RCDH) to learn effective binary codes for scalable multi-label image retrieval by listwise supervision. With deep convolutional neural networks, we can implement complex transformations on high-dimension raw data to obtain powerful representations. For the supervised learning with multiple labels, 
we rank the common class numbers between one image sample and other samples in the training dataset to build a rank list. 
By aligning the similarity between the semantic rank list in the original space and the distance rank list in the hamming space, the hamming distances between hashing codes can better reflect the semantic similarity of two images. We also design a powerful loss function to penalize the samples of which the similarities in two spaces are mismatched. Different from the multi-label hashing methods based on pairwise and triplet loss, our listwise objective can provide more global supervision information. To further exploit the semantic information provided by the multiple class labels, we extend the model to C-RCDH by including another two new losses into the objective function: a multi-label classification loss and a multi-label clustering loss. The multi-label classification term minimizes the multi-label softmax cross entropy loss to enhance the discriminative capacity of learned features. The multi-label clustering term is an extension of the center loss proposed for face recognition~\cite{wen2016discriminative}. This term reduces the distances between samples and the multiple class centers that the samples are associated with. 
As a result, we are able to exploit both the similarity reflected by common class numbers and the exact semantic information provided by the common labels. Therefore the model trained with the above loss functions can produce discriminative and polymerized hashing codes with appropriate neighborhood structure and thus strong retrieval ability. Experiments over three public multi-label image datasets, IAPRTC12, MIRFLICKR-25K and NUS-WIDE, show our C-RCDH method can provide state-of-the-art retrieval performance.

This work is the extension of our previous work~\cite{8486592} which has won the ICME 2018 Platinum Best Paper Award. The major extensions are in the following aspects:
\begin{itemize}
\item We modify the penalty functions in the term of rank-consistency loss. The intervals in hamming space are determined by an efficient way with an easier implementation. Moreover, we normalize the penalty distances by the length of hashing codes, so that the supervision signals will not disappear when the codes are long. 
\item We propose a multi-label clustering loss to enhance the manipulation on the distribution of the extracted features in the hamming space. With the assistant of the multi-label classification loss, the hashing codes have more powerful discriminative and polymerized capacity, which is important to achieve higher performance of multi-label image retrieval. 
\item We conduct extensive experiments on three widely used multi-label datasets, IAPRTC12, MIRFLICKR-25K and NUS-WIDE, in which NUS-WIDE is a larger-scale dataset with much more images and labels than the other two. Hence the state-of-the-art experimental results not only prove the effectiveness of the proposed method on relatively small dataset, but also show the superior performance on very large dataset. 
\end{itemize}

The rest of this paper is organized as follows: Section II discusses the related work of hashing methods. Our proposed method is described in detail in Section III. We introduce the experimental settings and present the experimental results on three datasets in Section IV. Finally, the last section is the conclusion of this paper.

\section{Related work}

Learning to rank~\cite{liu2011learning,cao2007learning,burges2005learning,xia2008listwise,li2011learning,severyn2015learning} has been an abstractive research area in information retrieval, natural language processing and data mining. It aims to solve ranking problems by building retrieval functions automatically from training data, and has been widely adopted in tasks such as document retrieval, answer searching, content-based image retrieval (CBIR), etc. In the field of CBIR, hashing is able to learn hashing functions with excellent speed and memory efficiency, and has been regarded as a prevailing solution to the well-known problem of approximate nearest neighbor (ANN) search. 
There are two categories in previous hashing methods. The first one is \emph{data-independent} methods, including Locality Sensitive Hashing(LSH)~\cite{andoni2006near} and its discriminative and kernelized extensions~\cite{kulis2009kernelized,jain2008fast}. These methods only utilize random projection and do not consider data distribution. Hence such methods are not powerful enough. Additionally, the precision relies heavily on the length of hashing codes. 

The second category is \emph{data-dependent} hashing methods and many efforts have been devoted to this field by now. These hashing methods can be further divided into three categories: unsupervised methods, semi-supervised methods and supervised methods. In terms of unsupervised methods, 
Gong \emph{et al.}~\cite{gong2013iterative} proposed to find a rotation which can minimize the quantization error in the mapping procedure by Iterative Quantization (ITQ). 
Weiss \emph{et al.}~\cite{weiss2009spectral} presented a method of Spectral Hashing (SH) and obtained the solutions by the eigenvectors of graph Laplacian after relaxing the objective to another problem. 
Kong \emph{et al.}~\cite{kong2012isotropic} developed an isotropic hashing method (IsoHash) which tries to find projection functions to ensure the different projected dimensions have isotropic variances. 
Liu \emph{et al.}~\cite{NIPS2014_5332} preserved the structure of local neighborhoods in the discrete space by a graph-based hashing model. 

In the literature of learning to rank, there are three main categories of approaches in terms of supervision formulations: pointwise supervision, pairwise supervision and listwise supervision. On this basis, the ideas of embedding learning based on relative distance are also widely studied in other vision field including face recognition~\cite{schroff2015facenet}, metric learning~\cite{zheng2019hardness}, image classification~\cite{wang2014learning} and person re-identification~\cite{cheng2016person}. Similarly, such supervisions are also exploited to learn effective hashing functions. The methods utilizing the similarity information provided by the labels can be classified into semi-supervised hashing methods and supervised hashing methods. The development of large-scale image retrieval is further promoted by such methods.  
Kulis and Darrel~\cite{kulis2009learning} proposed a Binary Reconstructive Embedding (BRE) method by minimizing the reconstrction error between the Euclidean distance and the Hamming distance. 
Wang \emph{et al.}~\cite{wang2010semi} proposed a semi-supervised method to solve the issue that labeled images are limited and noisy by minimizing the empirical error of labeled data and maximizing the variance of hashing codes. 
Moreover, Iterative Quantization with canonical correlation analysis (ITQ-CCA)~\cite{gong2013iterative} utilized the supervision information of image data and achieved pioneering performance compared to ITQ.
Liu \emph{et al.}~\cite{liu2012supervised} proposed a kernel-based hashing method which minimizes the hamming distances of similar pairs and maximizes the distances of dissimilar pairs. 
Li \emph{et al.}~\cite{li2013learning} employed a large-margin learning framework to obtain the optimal hashing functions based on column generation. 
Lin \emph{et al.}~\cite{lin2015supervised} used graph cuts and boosted decision trees to give the solution to the problem of binary code learning and hashing function learning respectively. 
Shen \emph{et al.}~\cite{shen2015supervised} alleviated the issue of discrete constrains by solving a sub-problem of regularization. 

As deep neural networks become more and more prevalent and achieve state-of-the-art performance in many computer vision tasks, deep learning based hashing methods~\cite{Liong2016deep,chen2018deep,yuan2018relaxation} also become mainstream due to the excellent representation capacity compared to the conventional hand-crafted or shallow features. 
Liong \emph{et al.}~\cite{erin2015deep} proposed to exploit nonlinear projections on image samples by a deep neural network. 
Srivastava \emph{et al.}~\cite{srivastava2012multimodal} presented a deep Boltzmann Machine to solve the problem of multi-modal image representation. 
Xia \emph{et al.}~\cite{xia2014supervised} proposed a deep hashing method based on convolutional neural networks to obtain the results in two steps, where approximate hash codes and hash functions are learned respectively. 
Differently, Lai \emph{et al.}~\cite{Lai_2015_CVPR} proposed a "one-stage" deep hashing method which uses the triplet ranking loss~\cite{NIPS2012_4808} to supervise the learning of hashing codes. 
Lin \emph{et al.}~\cite{Lin_2015_CVPR_Workshops} utilized a point-wise method to learn hashing codes with a hidden layer which extracts the latent representations of class labels. 
Li \emph{et al.}~\cite{li2015feature} proposed a deep pairwise-supervised hashing method (DPSH) to simultaneously learn representations and hashing codes with pairwise labels. 
Inspired by DPSH~\cite{li2015feature}, Wang \emph{et al.}~\cite{wang2016deep} proposed a deep hashing method which maximizes the likelihood of triplet labels. 
Besides, Liu \emph{et al.}~\cite{Liu_2016_CVPR} trained the CNN model by simultaneously maximizing the discriminability of the outputs and imposing regularization to get binary codes. 
Li \emph{et al.}~\cite{li2017deep} assumed that the learned hashing codes could also be used for classification. Hence the framework is trained by both pairwise labels and classification information.

Here we summarize more hashing methods for multi-label image retrieval. Wu \emph{et al.}~\cite{wu2017deep} proposed the first deep hashing method based on pairwise labels for multi-label and large-scale image retrieval. 
Wang \emph{et al.}~\cite{wang2013learning} presented a ranking-based method (RSH) by leveraging rank triplets to supervise the learning of hashing functions based on linear projection.  
Zhao \emph{et al.}~\cite{zhao2015deep} developed a deep hashing method, which tries to maintain the multi-level semantic relationship among images and optimizes a surrogate ranking loss formulated by triplets. 
However, supervision information provided by triplets are limited. This may lead to some local minimum in the procedure of training. Meanwhile, the only way to obtain more global supervision is to sample more triplets for training. This operation brings cubic expansion of training data and thus expensive computation costs. 
Wang \emph{et al.}~\cite{wang2013order} presented a method named Order Preserving Hashing (OPH). This method maximizes the alignment of two similarity orders, one from the hamming space and the other from the original space. The problem is converted to multiple binary classification tasks which are produced by assigning original image samples to several sets divided in the hamming space. 
Different from the above method, we partition the categories according to the common labels in the original space and align the similarity order of the corresponding categories in the hamming space. 
Since the required length of hashing codes is generally larger than the number of common labels between two images, our strategy to partition the categories is more effective and efficient than OPH. 
Besides, instance-aware hashing methods have also emerged. Lai \emph{et al.}~\cite{lai2016instance} learned instance-aware representations for the image data organized in groups for multiple labels with the assistant of Spatial Pyramid Pooling (SPP) layer~\cite{he2015spatial}. Huang \emph{et al.}~\cite{huang2018object} proposed to focus on the objects instead of the background in an image by learning a binary mask map to identify the location of the objects. 

\begin{figure*}[t]
  \centering
  \centerline{\epsfig{figure=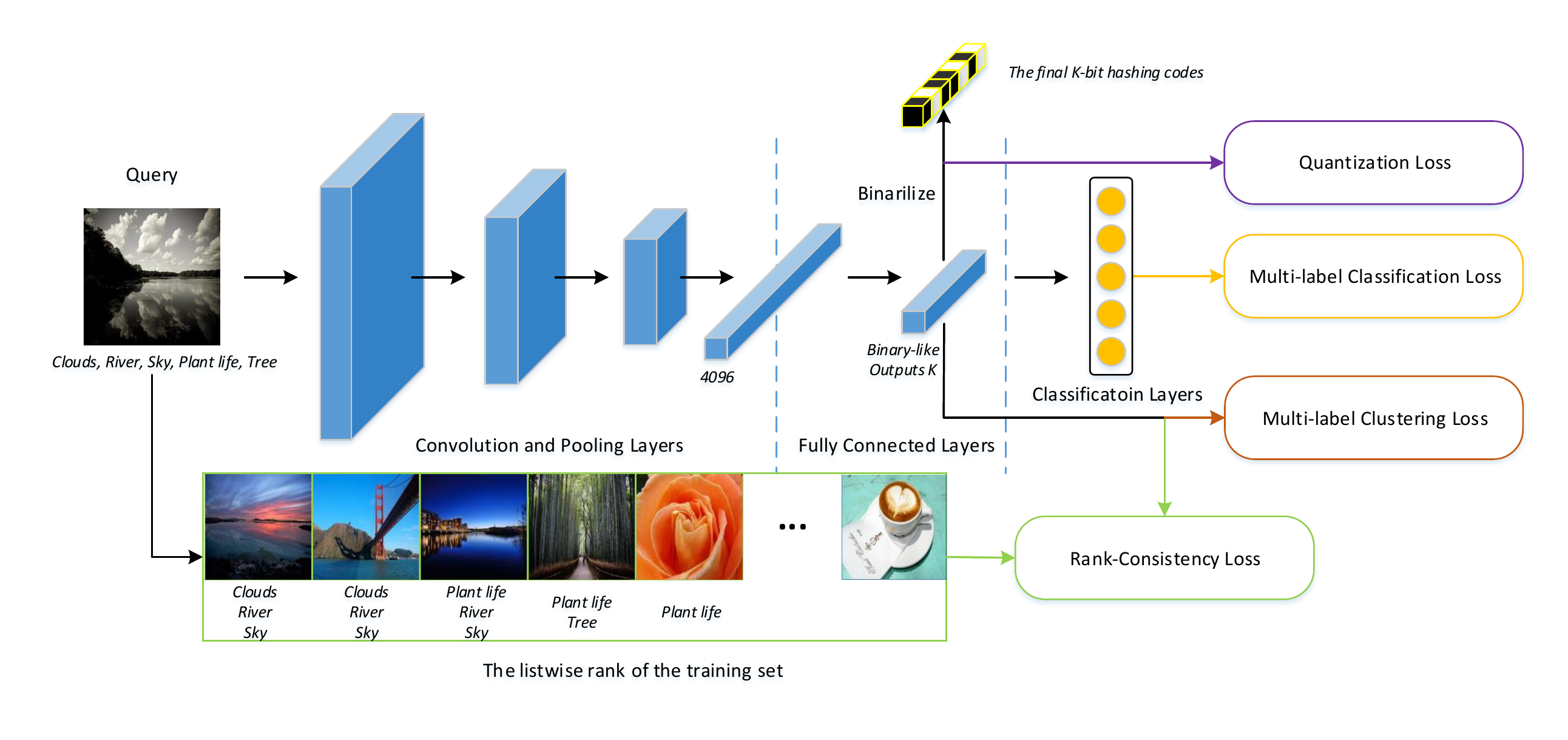,width=17cm}}
  \vspace{-5mm}
\caption{The framework of the proposed rank-consistency deep hashing method. We first extract real-valued embeddings using deep convolutional neural networks and get binary codes by the sign function. The deep model is trained by the rank-consistency loss, the multi-label classification and clustering losses and the quantization loss. In the retrieval procedure, the similarity between images is reflected by hamming distances. }
\label{fig:1}
\vspace{-3mm}
\end{figure*}

\section{Proposed Approach}

In this section, the details of the proposed method are described firstly. Then we present the formulations of each part in the objective function respectively. The overall view of our proposed approach is illustrated as Fig.~\ref{fig:1}. 

\subsection{Rank-Consistency Deep Hashing}

Firstly, we introduce the notation of the method. Given $\mathnormal{N}$ images $\mathcal{X}=\{\mathbf{x}_1, \mathbf{x}_2, \dots, \mathbf{x}_N\}, \mathbf{x}_n \in \mathbb{R}^{D}$ in the training set and their class labels $\mathcal{Y} = \{\mathbf{y}_1, \mathbf{y}_2, \dots, \mathbf{y}_N\}, \mathbf{y}_n \in \mathbb{R}^{C}$, where $y_{n,c}=1$ if image $n$ is associated with class label $c$ and $y_{n,c}=0$ otherwise. We aim to learn several hashing functions $\mathbf{h}(\mathbf{x})=(h_1(\mathbf{x}),h_2(\mathbf{x}),\dots,h_K(\mathbf{x}))^T$ to map the image samples into compact binary vectors of $K$ bits, $\{\mathbf{b}_1, \mathbf{b}_2, \dots, \mathbf{b}_N\}, \mathbf{b}_n \in \mathbb{H}^{K}$, where $\mathbb{H}^K$ denotes the $K$-dimension hamming space and $b_{n,k}=\{-1,1\}$ represents the $k$th bit in binary vector $n$. In this way, we can calculate the similarity of multi-label images efficiently. 

We aim at preserving the similarity relationship obtained in two spaces, the original space $\mathbb{R}^{D\times N}$ and the hamming space $\mathbb{H}^{K\times N}$. In the original space, the similarity order of an image can be easily obtained by sorting the common label numbers with other images in descending order, as $\mathbf{L}^O=(l^O_1, l^O_2, \dots, l^O_N)^T$. As for a binary vector in the hamming space, the order can be obtained similarly by sorting the hamming distances with other vectors, as $\mathbf{L}^H=(l^H_1, l^H_2, \dots, l^H_N)^T$. Effective hashing functions should keep $\mathbf{L}^O$ and $\mathbf{L}^H$ the same, that is $l^O_i=l^H_i, \forall i$. However, it is difficult and unnecessary to implement pointwise alignment. Thus we consider implementing listwise alignment by
partitioning $\mathbf{L}^O$ and $\mathbf{L}^H$ into some categories according to the common labels between images.

We can calculate the common label number between image sample $i$ and $j$ by $C_{ij}=\mathbf{y}_i^T\mathbf{y}_j$ in the original space. Assume there are $m_i$ different values in $\{C_{i,j}\}, j=1, \dots, N$, the original space can be divided to $m_i$ subsets as $\mathcal{S}^O_{i,1}, \mathcal{S}^O_{i,2}, \dots, \mathcal{S}^O_{i, m_i}$ by sorting $C_{i,j}$ in descending order. Consequently, $\mathcal{S}^O_{i,1}$ represents the image set nearest to the $i$ image and all the images in $\mathcal{S}^O_{i,j}$ have the same semantic distance to it. Besides, the hamming space can be separated accordingly. The hamming distance of $K$ bits can be partitioned to $m_i$ intervals, whose lower and upper bounds are denoted as $d^L_{i,j}$ and $d^U_{i, j}$ for $j=1,\dots,m_i$. We use $\mathcal{S}^H_{i,j}$ to represent the image samples whose hamming distances with $i$ are larger than $d^L_{i, j}$ and smaller than $d^U_{i,j}$. Then the key point of our method is to keep $\mathcal{S}^O_{i,j}$ and $\mathcal{S}^H_{i,j}$ consistent as depicted in Fig.~\ref{fig:21}. In practice, we can design different strategies to generate the intervals and the details will be presented below. 
The position and length of each interval can be determined according to the basic rule that large common label number means small hamming distance. 
Then we get the lower and upper bounds of the intervals. Thus, our method attempts to learn such hashing functions that the computed hamming distances between the $i$th image and the images in $\mathcal{S}^O_{i, j}$ are constrained into the interval $(d^L_{i,j},d^U_{i,j})$. 

For the above purpose, we penalize the mismatching images by the following loss functions:
\begin{eqnarray}
\mathrm{min} \mathcal{J}_r &=& \sum_{i=1}^{N}L_r(i) = \sum_{i=1}^{N}\sum_{j=1}^{N}L_r(i;j) \nonumber\\
~ &=& \sum_{i=1}^{N}\sum_{j=1}^{m_i}\sum_{k\in \mathcal{S}^O_{i,j}}f_{i,j}(D^H_{i,k}),  
\end{eqnarray}
where $L_r(i)$ is the rank loss of the $i$th image and $D^H_{i,k}$ is the hamming distance between the $i$th image and one image $k$ in the $j$th subset. $f_{i,j}(D^H_{i,k})$ computes the loss with respect to the hamming distance. Note that all image samples have the same common label numbers with $i$ in $\mathcal{S}^O_{i,j}$, and $N=\sum_{j=1}^{m_i}|\mathcal{S}^O_{i,j}|, \forall i$. Then what we need to do is specify the formulation of $f_{i,j}(\cdot)$. In fact, we aim to constrain $D^H_{i,k}\in (d^L_{i,j},d^U_{i,j})$. Therefore $f_{i,j}(D_{i,k})$ is defined as follows: 
\begin{eqnarray}
f_{i,j}(D^H_{i,k}) &=&  \mathrm{sign}(d^L_{i,j}-D^H_{i,k}) + \mathrm{sign}(D^H_{i,k}-d^U_{i,j}), 
\end{eqnarray}
in which $\mathrm{sign}(\cdot)$ is the sign function. Therefore, $f(\cdot)$ has the ability to evaluate whether the hamming distances of image samples are in proper intervals by penalizing the data points outside the corresponding interval. 

\textbf{Relaxation}: It can be observed that computing derivative for $\mathrm{sign}(\cdot)$ is intractable in practice. Following~\cite{wang2016deep}, we relax the discrete function and use a continuous function to alleviate this problem. Different from most methods which employ Sigmoid function $\sigma(z)=\frac{1}{1+e^{- z}}$ as the relaxing choice, 
we combine the logarithm function in the loss function as the modification. Then the formulation becomes $\phi(z)=-\mathrm{log}\sigma(-z)=\mathrm{log}(1+e^{ z})$. It has two significant advantages: 
\begin{itemize}
\item $\phi(z)$ is close to 0 when $z$ is less than 0. Otherwise $\phi(z)$ rises according to the increase of $z$. Thus the  penalty value depends on $z$, rather than a fixed value 1. 
\item $\phi'(z)=\frac{e^z}{1+e^{ z}}$ is the derivative of $\phi(z)$. When $z>0$, this value is approximately equal to 1. Hence the CNN model may converge faster using such loss function. 
\end{itemize}

We calculate the hamming distance between two vectors by 
\begin{eqnarray}
D^H_{i,k}=\frac{K-\mathbf{b}_i^T\mathbf{b}_k}{2}.
\end{eqnarray}
This step is also discrete. Following~\cite{wang2016deep}, we use the model output, $\mathbf{u}\in \mathbb{R}^{K}$, instead of $\mathbf{b}\in \{-1,1\}^K$ for relaxation: 
\begin{eqnarray}
\widetilde{D}^H_{i,k}=\frac{K-\mathbf{u}_i^T\mathbf{u}_k}{2}. 
\end{eqnarray}
Then the binary codes $\mathbf{b}$ are obtained by applying the sign function on $\mathbf{u}$ as $\mathbf{b}=\mathrm{sign}(\mathbf{u})$. Then we have the formulation of $\mathcal{J}_r$ for C-RCDH based on the above analysis, 

\begin{eqnarray}
\mathcal{J}_r = \sum_{i=1}^{N}\sum_{j=1}^{m_i}\sum_{k\in S^O_{i,j}}-\mathrm{log}\sigma \Bigg(\frac{\gamma}{K} \Big(\frac{K-\mathbf{u}_i^T\mathbf{u}_k}{2}-d^L_{i,j}\Big) \Bigg) \nonumber \\
 -\mathrm{log}\sigma \Bigg(\frac{\gamma}{K} \Big(d^U_{i,j}-\frac{K-\mathbf{u}_i^T\mathbf{u}_k}{2} \Big)\Bigg),  \label{eq:2}
\end{eqnarray}
where $\gamma/K$ is an important normalization factor. $K$ and $\gamma$ are the length of hashing codes and the parameter to balance the scale of penalties, respectively. The reason why we need this factor can be demonstrated as follows. 
When $z$ is a negative number with large absolute values, the derivative value $\phi'(z)$ will be small. In this case, the issue of gradient disappearance arises, which may make the neural networks lose the learning ability via back propagation. This issue will affect especially the retrieval performance of long hashing codes. However, with the afore-mentioned normalization technique, the derivative function $\phi'(z)$ will have large absolute values since the values of $z$ are restricted close to $0$. The distance terms $(K-\mathbf{u}_i^T\mathbf{u}_k)/2-d^L_{i,j}$ and $d^U_{i,j}-(K-\mathbf{u}_i^T\mathbf{u}_k)/2$ are positively correlated with the length of hashing codes $K$, so we just use $\gamma/K$ as the normalization factor. 

\label{sec:interval}
\textbf{Determination of Intervals}: If we denote the $m_i$ different numbers of common labels which are listed in descending order as $\{p_{i,j}\}, j=1, \dots, m_i$, we know the maximum value of the list $\{p_{i,j}\}$ is $p_{i,1}$ while the minimum one is $p_{i,m_i}$. In fact, for sample $i$, we only consider the relative similarity differences of the original space. Hence we determine the interval lengths by the relative value $\Delta p_i = p_{i,1}-p_{i,m_i}$, instead of the absolute value $\{p_{i,j}\}$. We denote a step length as $s=K/(\Delta p_i+2)$ and then, the lower and upper bounds can be obtained by $d_{i,j}^L=s\cdot (p_{i,1}-p_{i,j})$ and $d_{i,j}^H=s\cdot (p_{i,1}-p_{i,j}+2)$. We see this formulation also satisfies $d_{i,1}^L=0$ and $d^H_{i,m_i}=K$. Moreover, in this case, the interval separation is no longer affected by the number of common label numbers, $m_i$, but is only determined by the relative similarity differences, $p_{i,1}-p_{i,m_i}$ and $p_{i,1}-p_{i,j}$. The method to determine intervals is illustrated in Fig.~\ref{fig:21} (b). 

Note that for $x_i$ and $x_j$, the two rank lists of $x_i$ and $x_j$ can both determine the intervals where $D^H_{i,j}$ should be restricted in, i.e., $(d^L_{i,j}, d^H_{i,j})$ may be different from $(d^L_{j,i}, d^H_{j,i})$. While the two intervals may be different, the objective is able to automatically adjust the hamming distance $D^H_{i,j}$ to an appropriate value. For example, the hamming distance may be in the overlap region of the two intervals if they have overlap. Otherwise, $D^H_{i,j}$ may be optimized to lie in the  region between two intervals if they have no overlap. In both cases, our rank-consistency loss is able to guarantee a reasonable optimum.

\subsection{Multi-Label Classification and Clustering Losses}
\textbf{Multi-Label Classification Loss}: However, so far we have only used the number of common labels between images. We cannot fully exploit what common labels these images have by such supervision. Hence we need to provide this term of information through a layer which can be treated as a multi-classifier. 
Different from most multi-label classification methods which utilize the sigmoid cross-entropy loss,  
we propose a new multi-label softmax cross-entropy loss for the reason that we do not need obtain the exact labels that one image has. We just intend to maximize the responding magnitudes of the corresponding classes. Therefore the previous single-label softmax loss is modified to a multi-label extension so as to enhance the responses of multiple classes. Considering the conciseness of the formulation and its derivative, this term is defined as~(\ref{eq:1}). $\mathbf{W}\in \mathbb{R}^{C\times K}$ and $\mathbf{v} \in \mathbb{R}^{C\times 1}$ are the projection matrix and bias vector respectively. $\mathbf{1}$ is a vector with $C$ ones. 
\begin{eqnarray}
\label{eq:1} \mathcal{J}_{cla} &=& -\sum_{i=1}^N \mathrm{log}\left(\frac{\mathrm{exp}\left[\mathbf{y}_i^T(\mathbf{W}\mathbf{u}_i+\mathbf{v})\right]}{\mathbf{1}^T\mathrm{exp}(\mathbf{W}\mathbf{u}_i+\mathbf{v})}\right). 
\end{eqnarray}

\begin{figure*}[htbp]
\centering
\subfigure[]{
\begin{minipage}[t]{0.33\linewidth}
\centering
\includegraphics[width=6cm]{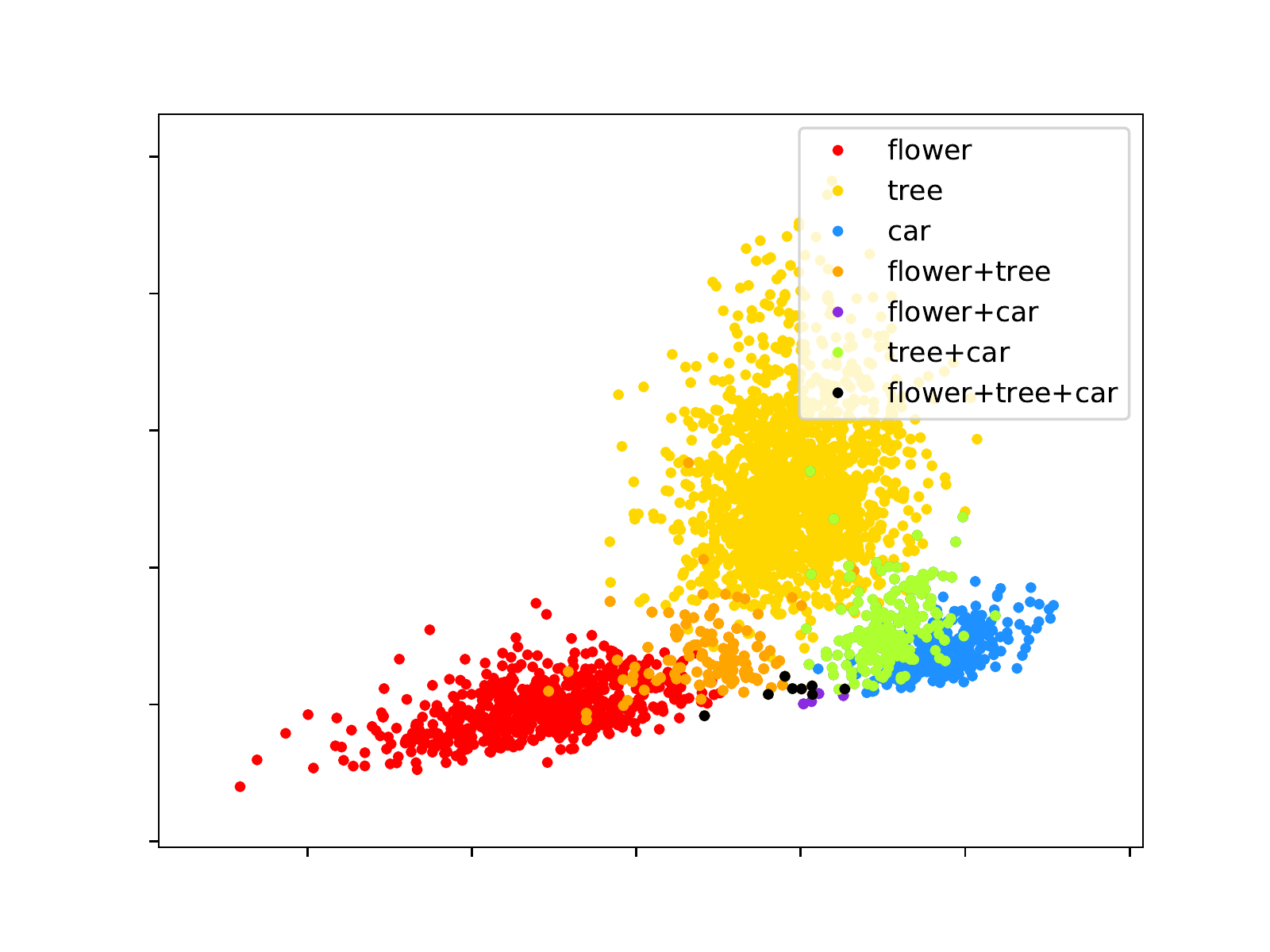} \\
\vspace{-3mm}
\includegraphics[width=6cm]{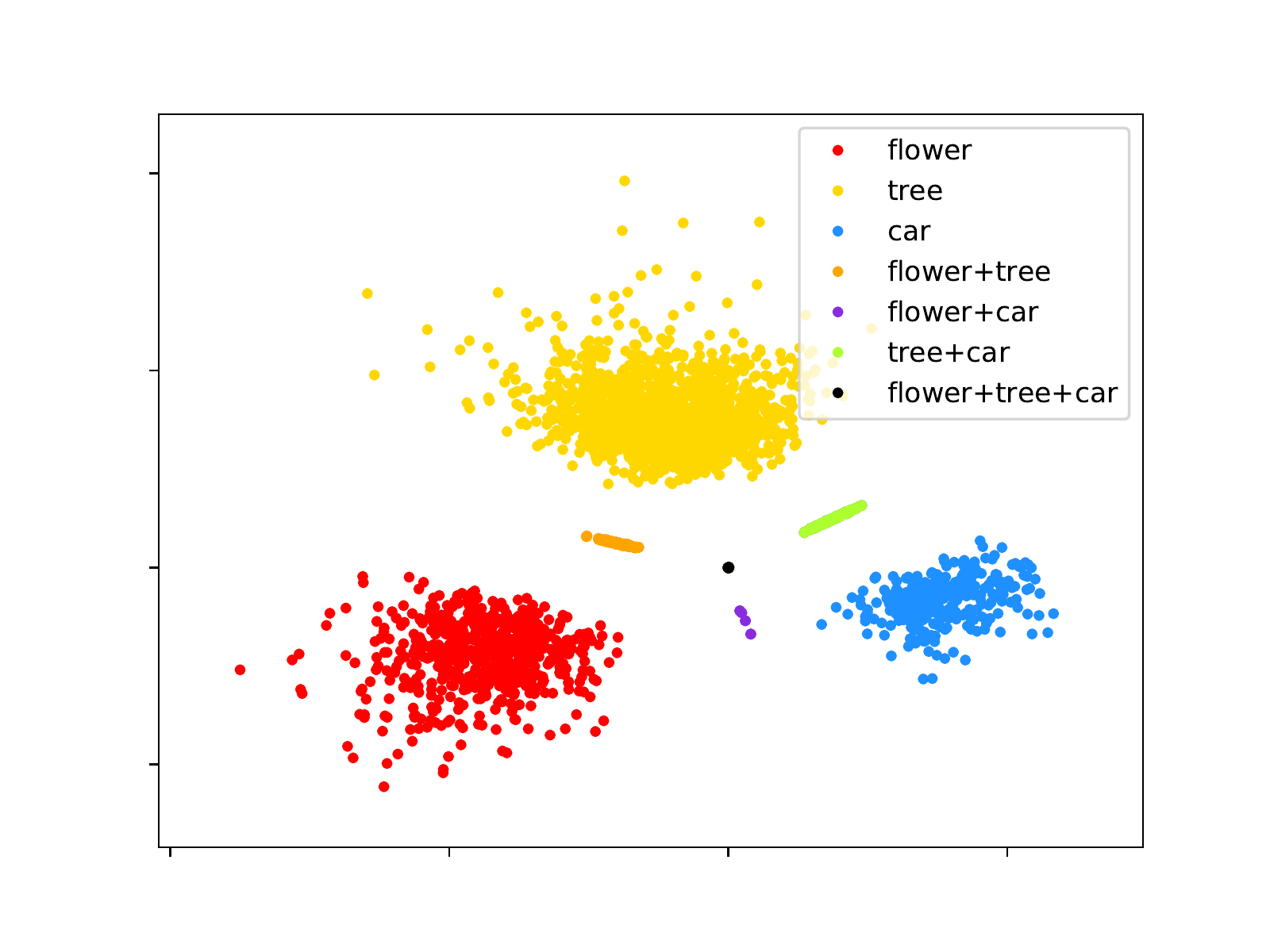}
\end{minipage}%
}%
\subfigure[]{
\begin{minipage}[t]{0.33\linewidth}
\centering
\includegraphics[width=6cm]{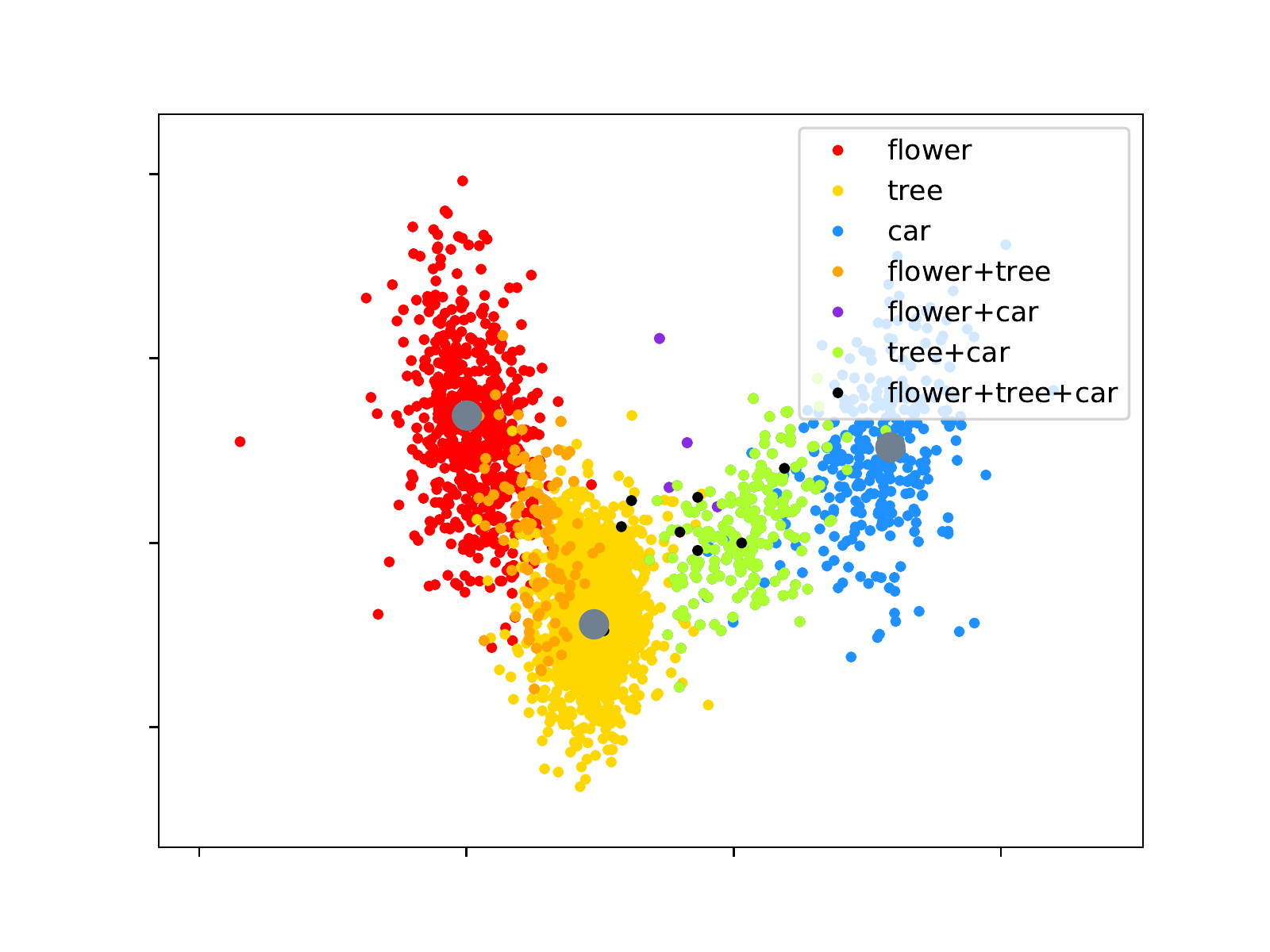} \\
\vspace{-3mm}
\includegraphics[width=6cm]{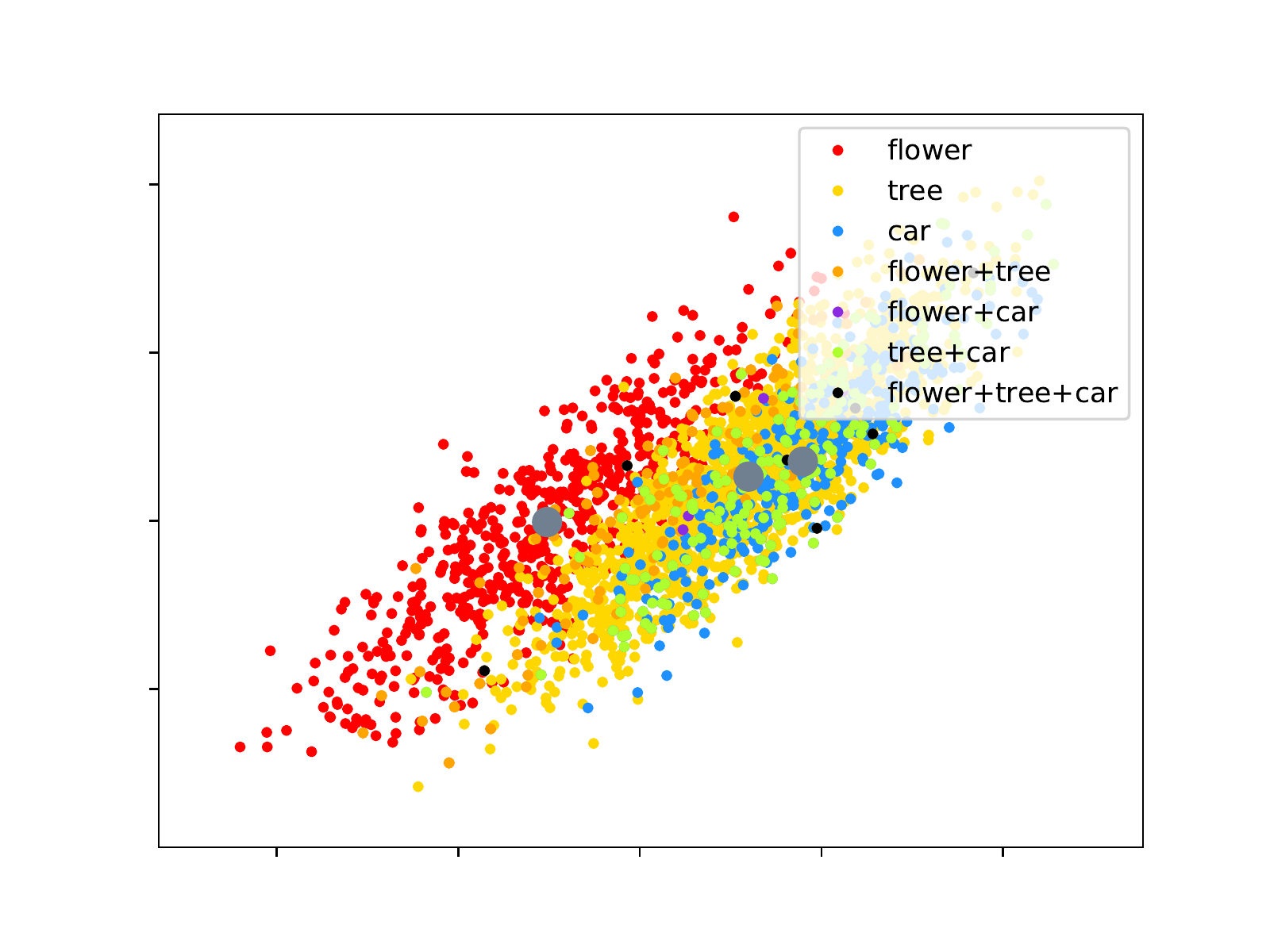}
\end{minipage}%
}%
\subfigure[]{
\begin{minipage}[t]{0.33\linewidth}
\centering
\includegraphics[width=6cm]{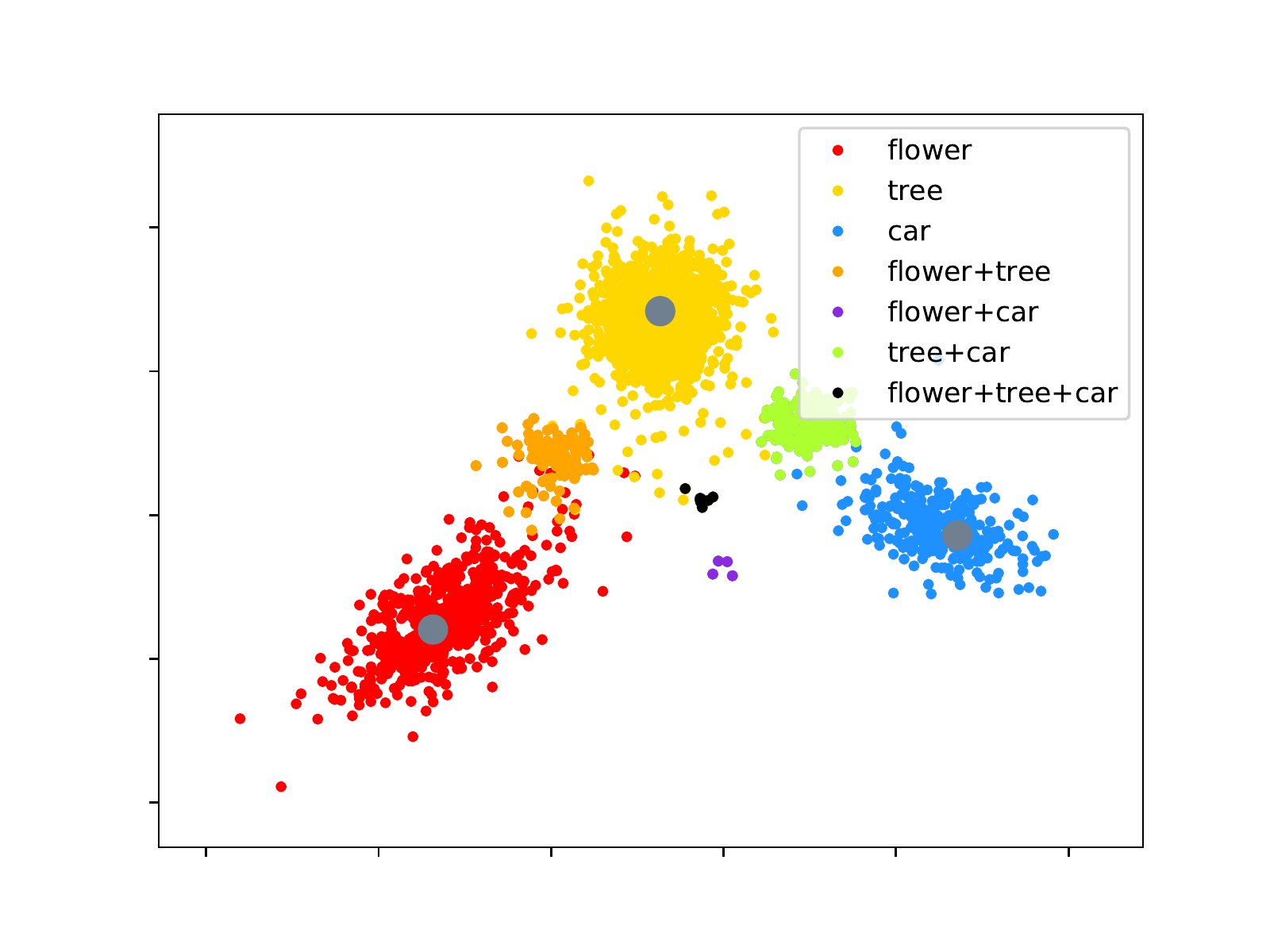} \\
\vspace{-3mm}
\includegraphics[width=6cm]{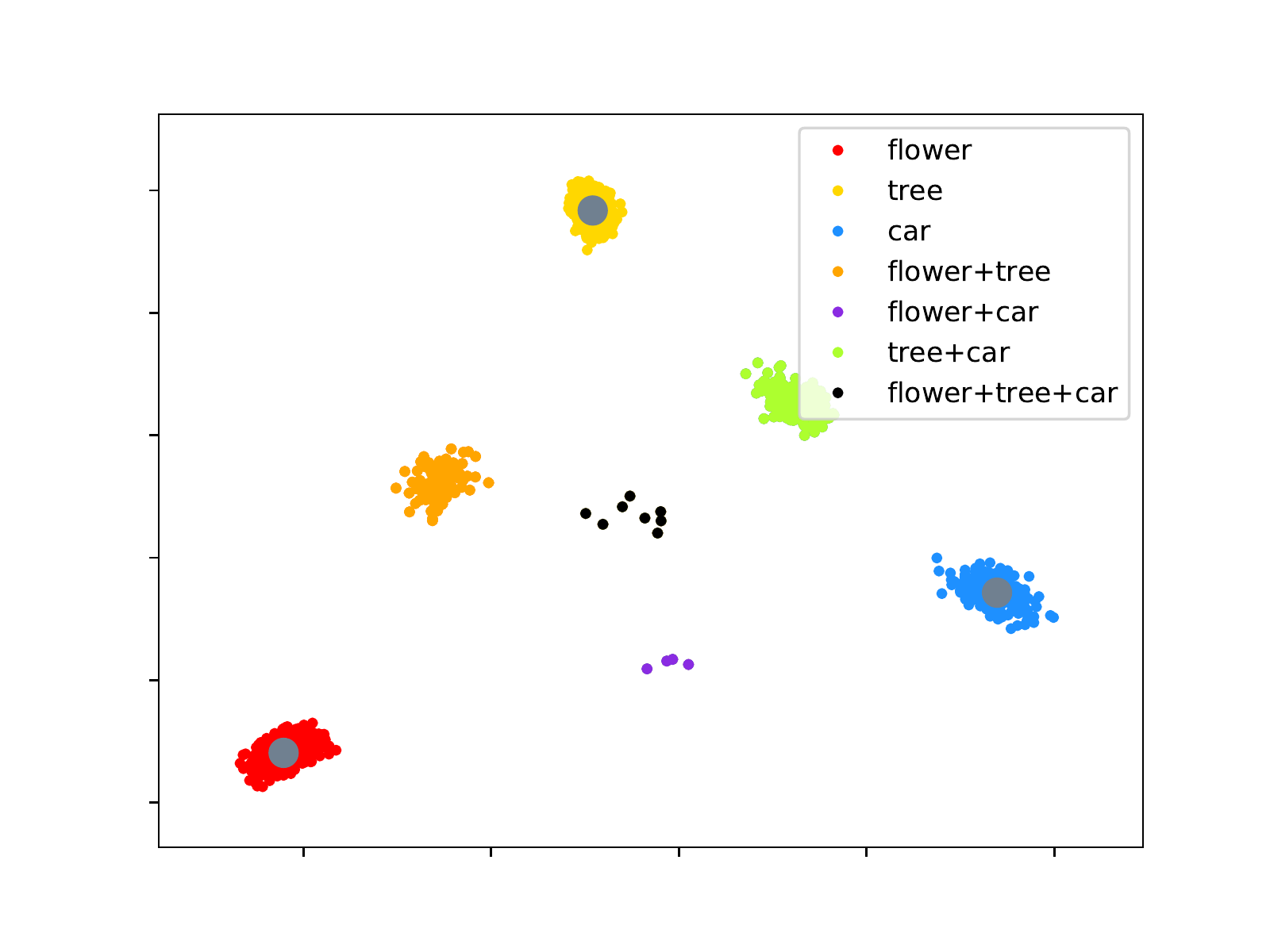}
\end{minipage}
}
\centering
\caption{Distribution of learned 2D features by deep neural networks with different objective functions: (a) only the multi-label classification loss, (b) only the multi-label clustering loss, (c) both the multi-label classification loss and the multi-label clustering loss. Samples are from MIRFLICKR-25K and associated with three labels, flower, tree and car. The distributions in the first row are before the convergences while those in the second row are near the convergences. The gray dots represent the centers of three classes. Best viewed in color. }
\label{fig:visual}
\vspace{-3mm}
\end{figure*}

\textbf{Multi-Label Clustering Loss}: In order to further improve the discriminative ability of hashing codes, we propose a novel multi-label clustering loss to reduce the hamming distances between the samples with the same labels. Different from the center loss proposed in~\cite{wen2016discriminative} for the task of face recognition, we do not have clear definitions for intra-class variations and inter-class distances of the extracted features in multi-label hashing. The hamming distance between two samples should be reduced as long as they have common labels. Moreover, when two samples have more common labels, two hashing vectors should be closer. Hence we just reduce the distances between samples and their related center vectors to construct polymerized neighborhood structures. We assign a center vector for each class, as $\mathbf{C} = \{\mathbf{c}_1, \mathbf{c}_2, \dots, \mathbf{c}_C\}, \mathbf{c}_t \in \mathbb{R}^K$, where $C$ is the total number of class labels and $K$ is the length of hashing codes. For an image sample $i$, the formulation of the multi-label clustering term should reduce the distance between $\mathbf{u}_i$ and $\mathbf{c}_t$ once $i$ has the $t$th class label. If we use Euclidean distance as the objective function, we have the following formulation: 
\begin{eqnarray}
\mathcal{J}_{clu} = \frac{1}{2} \sum_{i=1}^{N}\frac{\sum_{t=1}^{C} y_{i,t} \cdot \|\mathbf{u}_i-\mathbf{c}_t\|^2 }{\sum_{t=1}^C y_{i,t}+\epsilon},   \label{eq:cen}
\end{eqnarray}
where $y_{i,t}=1$ if sample $i$ has label $t$ and $y_{i,t}=0$ otherwise, and $\epsilon$ is a very small number to avoid the invalid calculation when a sample $i$ is not associated with any label. Therefore, the more class labels $i$ is associated with, the more terms of constraints there are on the feature of $i$. However, each term is assigned a weight of $1/(\sum_{t=1}^C y_{i,t}+\epsilon)$, which can control the updating scale of $\mathbf{u}_i$ and keep the supervision in balance. 

Then the center vector $\mathbf{c}_t$ represents the center of all the feature vectors with the label $t$. Hence $\mathbf{c}_t$ should be calculated by the average of feature vectors of all the samples in whose label vector $y_{i,t}$ is $1$. Following~\cite{wen2016discriminative}, we also update the centers vectors based on mini-batch in each iteration due to the intractability of calculating the centers by the entire training dataset. Hence the update function of the $t$th label is formulated as:
\begin{eqnarray}
\Delta \mathbf{c}_t = \frac{\sum_{i=1}^{m} y_{i,t} \cdot (\mathbf{c}_t-\mathbf{u}_i) }{\sum_{i=1}^{m} y_{i,t}+\epsilon},   \label{eq:deltac}
\end{eqnarray}
where $m$ is the batch size in the training process. With such supervision, the feature of a sample should be close to the centers of all the labels the sample is associated with. On the one hand, if we update feature vectors while fixing center vectors according to the multi-label clustering term, the samples with same labels will have small hamming distances in the hamming space. On the other hand, if we update center vectors by calculating the averages of the corresponding features, the center vectors of related labels tend to cluster close since some samples may have these labels simultaneously with a high probability. Therefore combined with the aforementioned rank-consistency term which can constrain the hamming distances in appropriate ranges, the multi-label classification term and this multi-label clustering term can help arrange the distribution of the hamming space accordingly to fully exploit the category information in the multiple labels. 

\label{sec:necessity}
\textbf{Necessity of Structure Constrains}: Here we discuss the necessity of the discriminative supervision, multi-label classification loss and the polymerized supervision, multi-label clustering loss. If there is merely a rank-consistency loss, the distances between features may be constrained. However, samples from different categories are located in a cluttered distribution since the supervision only utilizes the number of common labels between two samples, but not what exactly the labels are. In this case, the supervision ability may be limited since there is not a holistic geometric structure for the learned features. Therefore, the supervision on the neighborhood structure is desired for better performance. In order to show the effectiveness of the multi-label classification loss and the multi-label clustering loss, we implemented a toy experiment on learning 2D features for the images with three labels and their permutations as depicted in Fig.~\ref{fig:visual}. These samples are trained by different objective functions: 
\begin{itemize}
\item When the features are learned by the multi-label classification loss, linear boundaries between different categories are clear. However, this distribution obviously cannot meet the requirement that samples with more common labels should have smaller distances. Therefore, multi-label classification loss can only promote the learning to some extent. 
\item From the formulation of the multi-label clustering loss, it can be inferred that as long as the two samples have common labels, their features will be closer and closer. Meanwhile, once there are samples associated with two labels simultaneously, the center vectors of these two labels will also be closer gradually. Finally, all the related samples and center vectors will gather together, which is not what we need. 
\item When two terms are combined for the training, the features not only gain the discriminative ability, but also have an appropriate neighborhood structure with the assistant of the polymerized supervision. On the one hand, the features from different categories can be easily classified by decision boundaries. On the other hand, all the features cluster around the corresponding centers. Consequently, the distance between features is consistent with the level of semantic similarity. The retrieval performance can also benefit a lot from such neighborhood structure. 
\end{itemize}

\subsection{Overall Objective Function} 

In order to guarantee the rationality of the above relaxation, from $\mathbf{b}$ to $\mathbf{u}$, we need another term of quantization loss as below, 
\begin{eqnarray}
\mathcal{J}_q = \sum_{i=1}^{N}\|\mathbf{b}_i-\mathbf{u}_i\|^2 \label{eq:jq}. 
\end{eqnarray}
This can be regarded as a regularizer, which makes the network extract features with real-value elements around $1$ or $-1$ without losing the ability to discriminate between different class labels. Finally, we have the overall objective function, 
\begin{eqnarray}
\mathrm{min} \mathcal{J} = \mathcal{J}_r+\lambda_{cla} \mathcal{J}_{cla}+\lambda_{clu} \mathcal{J}_{clu} + \lambda_q \mathcal{J}_q, 
\end{eqnarray}
where $\lambda_{cla}$, $\lambda_{clu}$ and $\lambda_q$ are the parameters to weight multiple losses. It costs a lot to rank all the images in the training set in practice. Hence we may just choose the first $N_r$ samples from the order list for computational efficiency. 

\subsection{Gradient Descent}
The stochastic gradient descent algorithm is used to optimize the parameters of the deep model. Let us define:
\begin{eqnarray}
A_{i,k} = \frac{\gamma}{K} \Big(\frac{K-\mathbf{u}^T_i\mathbf{u}_k}{2}-d^L_{i,j}\Big), \\ 
B_{i,k} = \frac{\gamma}{K} \Big(d^U_{i,j}-\frac{K-\mathbf{u}^T_i\mathbf{u}_k}{2}\Big),
\end{eqnarray}
in which $k$ is an image in the $j$th subset of the $i$th image, $S^O_{i,j}$. Therefore, the derivatives of $\mathcal{J}_r$~(\ref{eq:2}), $\mathcal{J}_{cla}$~(\ref{eq:1}) and $\mathcal{J}_q$~(\ref{eq:jq}) with respect to output vector $\mathbf{u}_p$ can be computed by: 
\begin{eqnarray}
\frac{\partial \mathcal{J}_r}{\partial \mathbf{u}_p}&=&\sum_{i=1}^{N}\frac{\gamma\mathbf{u}_i}{2K}\left[\sigma(B_{i,p})-\sigma(A_{i,p})\right] \nonumber \\
&&+\sum_{k=1}^N\frac{\gamma\mathbf{u}_k}{2K}\left[\sigma(B_{p,k})-\sigma(A_{p,k})\right].  \\
\nonumber \\
\frac{\partial \mathcal{J}_{cla}}{\partial \mathbf{u}_p} &=& \mathbf{W}^T\left[\frac{\mathrm{exp}(\mathbf{W}\mathbf{u}_i+\mathbf{v})}{\mathbf{1}^T\mathrm{exp}(\mathbf{W}\mathbf{u}_i+\mathbf{v})}-\mathbf{y}_p\right]. \\
\nonumber \\
\frac{\partial \mathcal{J}_q}{\partial \mathbf{u}_p}&=&2(\mathbf{u}_p-\mathbf{b}_p).
\end{eqnarray}

For $\mathcal{J}_{clu}$~(\ref{eq:cen}), the gradients with respect to $\mathbf{u}_p$ are: 
\begin{eqnarray}
\frac{\partial \mathcal{J}_{clu}}{\partial \mathbf{u}_p}&=&\frac{\sum_{t=1}^{C} y_{p,t} \cdot (\mathbf{u}_p-\mathbf{c}_t) }{\sum_{t=1}^C y_{p,t}+\epsilon}.
\end{eqnarray}
Meanwhile, the updating equation for the center vectors $\mathbf{c}_t$ in the $j$th iteration are defined with~(\ref{eq:deltac}) as follows:
\begin{eqnarray}
\mathbf{c}^{j+1}_t&=&\mathbf{c}^j_t-\alpha \cdot \Delta\mathbf{c}^j_t,
\end{eqnarray}
where $\alpha$ is a parameter to control the updating step of center vectors. When $\alpha$ is too large, the center vectors may be unstable as a result of the frequent content changes in different mini-batches. When $\alpha$ is too small, the term of multi-label clustering loss may lead to a slow convergence since it may take a long time to approximate the center vectors accurately compared to the results calculated by the entire training dataset. 

\begin{figure}[t]
\centering
\subfigure[]{
\begin{minipage}[t]{\linewidth}
\centering
\includegraphics[width=8.5cm]{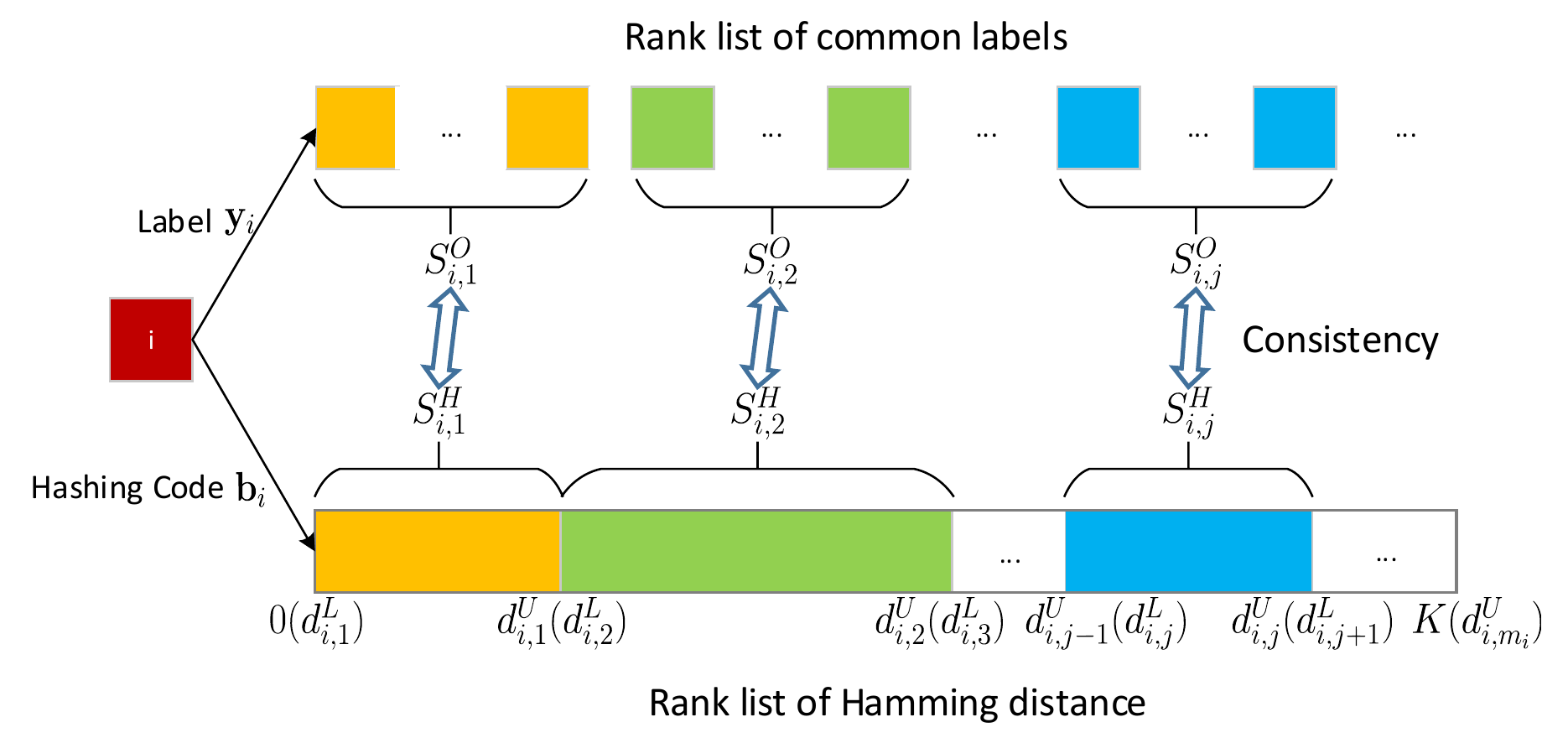}
\end{minipage}%
}%

\subfigure[]{
\begin{minipage}[t]{\linewidth}
\centering
\includegraphics[width=8.5cm]{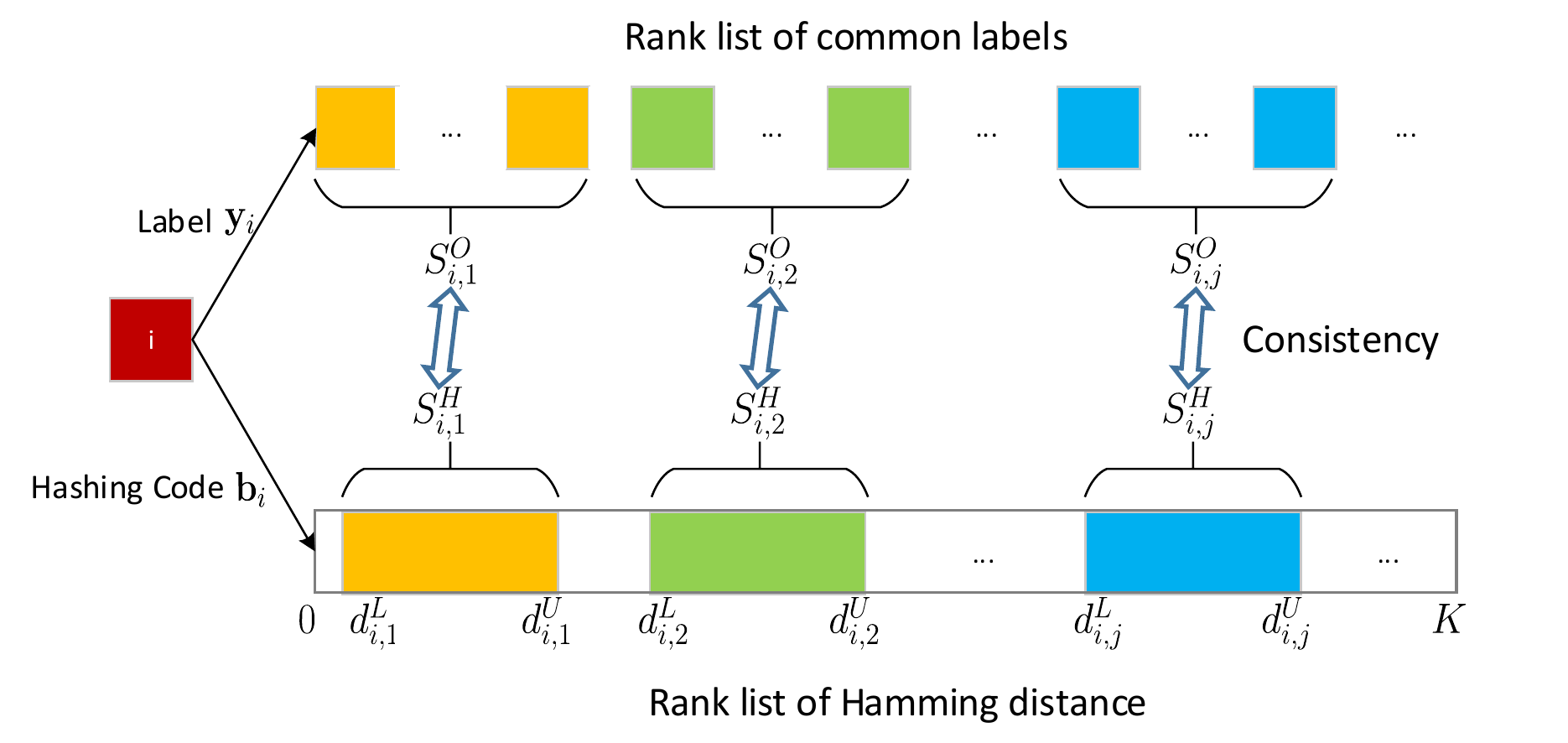}
\end{minipage}%
}%

\centering
\caption{We sort the common label numbers in descending order and divide the original space to $m_i$ subsets accordingly as  $S^O_{i,1}, S^O_{i,2}, \dots, S^O_{i, m_i}$. The rank-consistency is preserved when the hamming distances between the images in subset $S^O_{i, j}$ and image $i$ are in the partitioned interval $(d^L_{i,j},d^U_{i,j})$. In~\cite{8486592}, the whole hamming space is separated into $m_i$ intervals (a). In the C-RCDH method the extracted intervals have similar length  (b). }
\label{fig:21}
\vspace{-3mm}
\end{figure}

\begin{table*}[t]
\centering
\caption{Comparison of $NDCG@100$ and $ACG@100$ performance among conventional methods on the dataset of MIRFLICKR-25K.}
\begin{tabular}[width =7 in]{|c|c|c|c|c|c|c|c|c|c|c|}
\hline
 \multicolumn{1}{|c|}{} & \multicolumn{5}{c|}{$NDCG@100$} & \multicolumn{5}{c|}{$ACG@100$} \\ \hline
Method & 16 bits & 32 bits & 48 bits & 64 bits & 128 bits & 16 bits & 32 bits & 48 bits & 64 bits & 128 bits \\ \hline
PCA-H~\cite{wang2010semi} & 0.2566 & 0.2676 & 0.2711 & 0.2761 & 0.2797 & 1.9850 & 2.0015 & 2.0103 & 2.0336 & 2.0383 \\ \hline
PCA-ITQ~\cite{gong2013iterative} & 0.2814 & 0.3113 & 0.3209 & 0.3293 & 0.3391 & 2.1945 & 2.3318 & 2.3744 & 2.4109 & 2.4518 \\ \hline
SPLH~\cite{wang2010semi} & 0.2754 & 0.3081 & 0.3151 & 0.3211 & 0.3323 & 2.2009 & 2.3711 & 2.4084 & 2.4338 & 2.4882 \\ \hline
KSH~\cite{liu2012supervised} & 0.3033 & 0.3350 & 0.3498 & 0.3527 & 0.3584 & 2.3098 & 2.4360 & 2.5097 & 2.5126 & 2.5266 \\ \hline
CGH~\cite{li2013learning} & 0.3302 & 0.3592 & 0.3810 & 0.3872 & 0.3910 & 2.3981 & 2.5681 & 2.6710 & 2.7012 & 2.7213 \\ \hline
FH~\cite{lin2015supervised} & 0.2507 & 0.2813 & 0.2866 & 0.2960 & 0.3328 & 1.9401 & 2.1770 & 2.1560 & 2.2030 & 2.4019 \\ \hline
SDH~\cite{shen2015supervised} & 0.2711 & 0.2810 & 0.2836 & 0.2935 & 0.3186 & 2.2540 & 2.3458 & 2.3715 & 2.4114 & 2.4894 \\ \hline
CCA-ITQ~\cite{gong2013iterative} & 0.3516 & 0.3876 & 0.4019 & 0.4073 & 0.4167 & 2.5703 & 2.7002 & 2.7784 & 2.7928 & 2.8381 \\ \Xhline{1.2pt}
C-RCDH & \textbf{0.4786} & \textbf{0.5119} & \textbf{0.5154} & \textbf{0.5284} & \textbf{0.5406} & \textbf{3.1273} & \textbf{3.2499} & \textbf{3.2454} & \textbf{3.3197} & \textbf{3.3679} \\ \hline 
\end{tabular}
\label{table:result_mir_s}
\end{table*}

\begin{table*}[t]
\centering
\caption{Comparison of $NDCG@100$ and $ACG@100$ performance among conventional methods on the dataset of IAPRTC12.}
\begin{tabular}[width =7 in]{|c|c|c|c|c|c|c|c|c|c|c|}
\hline
 \multicolumn{1}{|c|}{} & \multicolumn{5}{c|}{$NDCG@100$} & \multicolumn{5}{c|}{$ACG@100$} \\ \hline
Method & 16 bits & 32 bits & 48 bits & 64 bits & 128 bits & 16 bits & 32 bits & 48 bits & 64 bits & 128 bits \\ \hline
PCA-H\cite{wang2010semi} & 0.1784 & 0.1898 & 0.1951 & 0.1958 & 0.2044 & 0.7951 & 0.7975 & 0.7947 & 0.7845 & 0.7882 \\ \hline
PCA-ITQ\cite{gong2013iterative} & 0.1947 & 0.2168 & 0.2323 & 0.2407 & 0.2575 & 0.8600 & 0.9176 & 0.9538 & 0.9788 & 1.0232 \\ \hline
SPLH\cite{wang2010semi} & 0.2154 & 0.2397 & 0.2404 & 0.2413 & 0.2431 & 0.9465 & 1.0225 & 1.0302 & 1.0360 & 1.0450 \\ \hline
KSH\cite{liu2012supervised} & 0.2240 & 0.2477 & 0.2624 & 0.2696 & 0.2789 & 1.0092 & 1.0721 & 1.1144 & 1.1302 & 1.1535 \\ \hline
CGH\cite{li2013learning} & 0.2377 & 0.2642 & 0.2878 & 0.2895 & 0.2934 & 0.9984 & 1.0147 & 1.0810 & 1.1596 & 1.1672 \\ \hline
FH\cite{lin2015supervised} & 0.1567 & 0.1931 & 0.2030 & 0.2122 & 0.2282 & 0.7914 & 0.9186 & 0.9660 & 0.9881 & 1.0489 \\ \hline
SDH\cite{shen2015supervised} & 0.1489 & 0.1859 & 0.2087 & 0.2327 & 0.2603 & 0.7351 & 0.8259 & 0.8971 & 0.9659 & 1.0424 \\ \hline
CCA-ITQ\cite{gong2013iterative} & 0.2475 & 0.2829 & 0.2967 & 0.3031 & 0.3148 & 1.0508 & 1.1261 & 1.1613 & 1.1728 & 1.2001 \\ \Xhline{1.2pt}
C-RCDH & \textbf{0.4388} & \textbf{0.4811} & \textbf{0.5050} & \textbf{0.5095} & \textbf{0.5369} & \textbf{1.1897} & \textbf{1.2804} & \textbf{1.3273} & \textbf{1.3477} & \textbf{1.4009}\\ \hline
\end{tabular}
\label{table:result_iap_s}
\end{table*}

\begin{table*}[t]
\centering
\caption{Comparison of $NDCG@100$ and $ACG@100$ performance among conventional methods on the dataset of NUS-WIDE.}
\begin{tabular}[width =7 in]{|c|c|c|c|c|c|c|c|c|c|c|}
\hline
 \multicolumn{1}{|c|}{} & \multicolumn{5}{c|}{$NDCG@100$} & \multicolumn{5}{c|}{$ACG@100$} \\ \hline
Method & 16 bits & 32 bits & 48 bits & 64 bits & 128 bits & 16 bits & 32 bits & 48 bits & 64 bits & 128 bits \\ \hline
PCA-H\cite{wang2010semi} & 0.3134 & 0.3401 & 0.3533 & 0.3637 & 0.3850 & 0.8864 & 0.9634 & 0.9997 & 1.0230 & 1.0774 \\ \hline
PCA-ITQ\cite{gong2013iterative} & 0.3407 & 0.3856 & 0.4000 & 0.4141 & 0.4329 & 0.9653 & 1.0767 & 1.1162 & 1.1510 & 1.1935 \\ \hline
SPLH\cite{wang2010semi} & 0.3472 & 0.3528 & 0.3519 & 0.3512 & 0.3483 & 0.9673 & 0.9852 & 0.9829 & 0.9826 & 0.9759 \\ \hline
KSH\cite{liu2012supervised} & 0.3473 & 0.3753 & 0.3809 & 0.3875 & 0.4002 & 0.9710 & 1.0523 & 1.0719 & 1.0943 & 1.1255 \\ \hline
CGH\cite{li2013learning} & 0.3629 & 0.4002 & 0.4180 & 0.4268 & 0.4464 & 0.9957 & 1.0963 & 1.1527 & 1.1831 & 1.2340 \\ \hline
FH\cite{lin2015supervised} & 0.2856 & 0.3091 & 0.3279 & 0.3479 & 0.3641 & 0.8068 & 0.8960 & 0.9440 & 1.0025 & 1.0442 \\ \hline
SDH\cite{shen2015supervised} & 0.2661 & 0.2989 & 0.3112 & 0.3173 & 0.3453 & 0.7999 & 0.8821 & 0.9075 & 0.9188 & 0.9907 \\ \hline
CCA-ITQ\cite{gong2013iterative} & 0.3892 & 0.4296 & 0.4492 & 0.4610 & 0.4803 & 1.1351 & 1.2332 & 1.2768 & 1.3087 & 1.3481 \\ \Xhline{1.2pt}
C-RCDH & \textbf{0.5223} & \textbf{0.5483} & \textbf{0.5655} & \textbf{0.5660} & \textbf{0.5744} & \textbf{1.3440} & \textbf{1.4078} & \textbf{1.4560} & \textbf{1.4506} & \textbf{1.4815}\\ \hline
\end{tabular}
\label{table:result_nus_s}
\end{table*}

\subsection{Discussion}

Here, we would like to discuss the difference of the interval determination between C-RCDH and RCDH~\cite{8486592}. First, we clarify the implementation details of the interval determination technique in RCDH. In order to determine the intervals for sample $i$, the strategy in RCDH is to partition the hamming distance of K into $m_i$ parts according to the ordered list of $m_i$ numbers of common labels. What we need to do is to determine how long each interval should be, since after that the lower and upper bounds $d^L_{i,j}$ and $d^H_{i,j}$ can be easily obtained. Considering that the interval length reflect the hamming distance and the relative similarity difference, the length of each interval is determined by the differences between two consecutive numbers in the ordered list of common labels. Specifically, we denote the $m_i$ different numbers of common labels which are listed in descending order, as $\{p_{i,j}\}, j=1, \dots, m_i$. The differences between two consecutive numbers can be calculated by $\Delta p_{i,j}=p_{i,j-1}-p_{i,j}$ accordingly. Note that we empirically define $\Delta p_{i,1}=1$ since we only care about the relative distance and we want the hamming distances of the most similar samples with $p_{i,1}$ common labels are constrained in a short interval. Then, the length of each interval can be obtained by $\Delta d_{i,j}=d^H_{i,j}-d^L_{i,j}=K\cdot \Delta p_{i,j}/\sum_{j=1}^{m_i} \Delta p_{i,j}$ and we get all the intervals once we apply $d^L_{i,1}=0$ and $d^H_{i,m_i}=K$. However, such strategy has little supervision for the samples which only have a small number of common label numbers with other samples, that is, $m_i$ is very small. In this case the hamming space of $K$ bits is separated into only few intervals and the length of each interval is relatively large. Consequently, the penalty signals may be too weak to provide powerful training. Hence we introduce a more effective strategy to determine the intervals, as described in Section~\ref{sec:interval}. The strategy is modified in two aspects: (1) It is unnecessary that the sum of the lengths of all the intervals is equal to $K$. (2) The length of each interval can be a relatively small number to keep the supervision intensity. The comparison of two strategies is shown in Fig.~\ref{fig:21}. 

\section{Experiments}

We implement extensive experiments on three public multi-label image datasets to evaluate the proposed C-RCDH method. The details are presented as below. 

\subsection{Implementation Details}

We use the VGG\footnote{http://www.vlfeat.org/matconvnet/pretrained/} network as the architecture of our model to extract features. The model has 5 convolutional layers and there are normalization and max-pooling layers after the first two convolutional layers. The model also contains 3 fully-connected layers whose parameters are learnable. The last layer can map the features of length 4096 to the required code length $K$. We initialize the parameters by the CNN model pretrained on the dataset of ImageNet ILSVRC-2012~\cite{russakovsky2015imagenet} using the objective function of multinomial logistic regression. 
For the inputs to the model, we rescale the images to $224\times224\times3$ instead of cropping them in order to maintain the complete semantic structure. Additionally, the ReLU activation layer is used in our model. In the training process, the learning rate and batch size are set to 0.0001 and 48 respectively. 
$\lambda_q$ is set to $50$ empirically according to the characteristics of the datasets. 
$\lambda_{cla}$ and $\lambda_{clu}$ are both set to $20$ in order to work better combined with the rank-consistency term.  Additionally, we select the ranking scale $N_r$ as 10000 for each multi-label dataset. For the updating of the center vectors, $\alpha$ is fixed to $0.5$ in our experiments. In the formulation of rank-consistency loss, $\gamma$ is set to 16 to provide proper supervision to the networks. 

\subsection{Datasets and Evaluation Metrics}

\begin{table}[t]
\centering
\caption{Comparison of $NDCG@100$ and $ACG@100$ performance among deep multi-label methods on the dataset of MIRFLICKR-25K.}
\begin{tabular}[width =3 in]{|c|c|c|c|c|}
\hline
\multicolumn{5}{|c|}{$NDCG@100$} \\ \hline
Method & 16 bits & 32 bits & 48 bits & 64 bits \\ \hline
DPSH\cite{li2015feature} & 0.350 & 0.315 & 0.325 & 0.328 \\ \hline
DTSH\cite{wang2016deep} & 0.358 & 0.382 & 0.390 & 0.396 \\ \hline
DSRH\cite{zhao2015deep} & 0.415 & 0.470 & 0.490 & 0.505  \\ \hline
ISDH\cite{zhang2018instance} & 0.377 & 0.396 & 0.401 & 0.406 \\ \hline 
DSPH\cite{yao2016deep} & 0.385 & 0.411 & 0.415 & 0.426 \\ \hline
MDSH\cite{liong2018multi} & 0.420 & 0.438 & 0.445 & 0.448 \\ \hline
RCDH\cite{8486592} & 0.438 & 0.481 & 0.483 & 0.488 \\ \Xhline{1.2pt}
C-RCDH & \textbf{0.479} & \textbf{0.512} & \textbf{0.515} & \textbf{0.528}  \\ \hline
\multicolumn{5}{|c|}{$ACG@100$} \\ \hline
Method & 16 bits & 32 bits & 48 bits & 64 bits \\ \hline
DPSH\cite{li2015feature} & 2.510 & 2.408 & 2.435 & 2.438 \\ \hline
DTSH\cite{wang2016deep} & 2.566 & 2.675 & 2.699 & 2.763 \\ \hline
DSRH\cite{zhao2015deep} & 2.798 & 2.978 & 3.091 & 3.152  \\ \hline
ISDH\cite{zhang2018instance} & 2.632 & 2.730 & 2.782 & 2.795 \\ \hline
DSPH\cite{yao2016deep} & 2.653 & 2.768 & 2.773 & 2.928 \\  \hline
MDSH\cite{liong2018multi} & 2.844 & 2.919 & 2.963 & 2.961 \\ \hline
RCDH\cite{8486592} & 2.919 & 3.091 & 3.098 & 3.115 \\ \Xhline{1.2pt}
C-RCDH & \textbf{3.127} & \textbf{3.250} & \textbf{3.245} & \textbf{3.320}  \\ \hline
\end{tabular}
\label{table:result_mir_d}
\end{table}
\begin{table}[t]
\centering
\caption{Comparison of $NDCG@100$ and $ACG@100$ performance among deep multi-label methods on the dataset of IAPRTC12.}
\begin{tabular}[width =1 in]{|c|c|c|c|c|}
\hline
\multicolumn{5}{|c|}{$NDCG@100$} \\ \hline
Method & 16 bits & 32 bits & 48 bits & 64 bits \\ \hline
DPSH\cite{li2015feature} & 0.293 & 0.315 & 0.325 & 0.328 \\ \hline
DTSH\cite{wang2016deep} & 0.340 & 0.391 & 0.405 & 0.423 \\ \hline
DSRH\cite{zhao2015deep} & 0.394 & 0.433 & 0.448 & 0.465 \\ \hline
ISDH\cite{zhang2018instance} & 0.298 & 0.299 & 0.300 & 0.302 \\ \hline
DSPH\cite{yao2016deep} & 0.414 & 0.419 & 0.423 & 0.429 \\ \hline
RCDH\cite{8486592} & 0.421 & 0.453 & 0.463 & 0.471 \\ \Xhline{1.2pt}
C-RCDH & \textbf{0.439} & \textbf{0.481} & \textbf{0.505} & \textbf{0.510}  \\ \hline
\multicolumn{5}{|c|}{$ACG@100$} \\ \hline
Method & 16 bits & 32 bits & 48 bits & 64 bits \\ \hline
DPSH\cite{li2015feature} & 1.045 & 1.061 & 1.075 & 1.079 \\ \hline
DTSH\cite{wang2016deep} & 1.087 & 1.114 & 1.128 & 1.159 \\ \hline
DSRH\cite{zhao2015deep} & 1.117 & 1.194 & 1.232 & 1.255 \\ \hline
ISDH\cite{zhang2018instance} & 0.872 & 0.874 & 0.873 & 0.882 \\ \hline
DSPH\cite{yao2016deep} & 1.123 & 1.141 & 1.146 & 1.158 \\  \hline
RCDH\cite{8486592} & 1.157 & 1.226 & 1.254 & 1.271 \\ \Xhline{1.2pt}
C-RCDH & \textbf{1.190} & \textbf{1.280} & \textbf{1.327} & \textbf{1.348} \\ \hline
\end{tabular}
\label{table:result_iap_d}
\end{table}
\begin{table}[t]
\centering
\caption{Comparison of $NDCG@100$ and $ACG@100$ performance among deep multi-label methods on the dataset of NUS-WIDE.}
\begin{tabular}[width =1 in]{|c|c|c|c|c|}
\hline
\multicolumn{5}{|c|}{$NDCG@100$} \\ \hline
Method & 16 bits & 32 bits & 48 bits & 64 bits \\ \hline
DPSH\cite{li2015feature} & 0.343 & 0.320 & 0.347 & 0.350 \\ \hline
DTSH\cite{wang2016deep} & 0.352 & 0.375 & 0.397 & 0.413 \\ \hline
DSRH\cite{zhao2015deep} & 0.410 & 0.470 & 0.500 & 0.520 \\ \hline
ISDH\cite{zhang2018instance} & 0.413 & 0.418 & 0.422 & 0.423 \\ \hline 
DSPH\cite{yao2016deep} & 0.424 & 0.427 & 0.435 & 0.440 \\ \hline 
MDSH\cite{liong2018multi} & 0.457 & 0.482 & 0.495 & 0.502 \\ \hline 
RCDH\cite{8486592} & 0.474 & 0.515 & 0.525 & 0.531 \\ \Xhline{1.2pt}
C-RCDH & \textbf{0.522} & \textbf{0.548} & \textbf{0.566} & \textbf{0.566} \\ \hline
\multicolumn{5}{|c|}{$ACG@100$} \\ \hline
Method & 16 bits & 32 bits & 48 bits & 64 bits \\ \hline
DPSH\cite{li2015feature} & 0.968 & 0.924 & 0.982 & 0.995 \\ \hline
DTSH\cite{wang2016deep} & 0.997 & 1.052 & 1.104 & 1.152 \\ \hline
DSRH\cite{zhao2015deep} & 1.161 & 1.322 & 1.385 & 1.429 \\ \hline
ISDH\cite{zhang2018instance} & 1.108 & 1.095 & 1.097 & 1.117 \\ \hline
DSPH\cite{yao2016deep} & 1.146 & 1.132 & 1.155 & 1.187 \\  \hline
MDSH\cite{liong2018multi} & 1.269 & 1.338 & 1.367 & 1.384 \\ \hline
RCDH\cite{8486592} & 1.278 & 1.324 & 1.351 & 1.384 \\ \Xhline{1.2pt}
C-RCDH & \textbf{1.344} & \textbf{1.408} & \textbf{1.456} & \textbf{1.451} \\ \hline
\end{tabular}
\label{table:result_nus_d}
\end{table}

\textbf{MIRFLICKR-25K Dataset}\footnote{http://press.liacs.nl/mirflickr/}: In this dataset, there are 25000 images in total which are annotated manually with 38 concepts. Among the annotations, 24 includes varied concepts of scenes such as night, sky and indoor, and concepts of objects, for example, food, people and tree. For the other 14 classes, the labeling is stricter. Images are labeled with such concepts only if they are salient to the images. We choose 2000 images randomly to test the model and use the remaining as the training and gallery set following the settings used in~\cite{zhao2015deep}. 

\textbf{IAPRTC12 Dataset}\footnote{http://www.imageclef.org/photodata}: This dataset contains 19627 images which reflect contemporary life in many aspects, for example, landscapes, cities, nature and people. There are 275 segmentation annotations for the images. The most frequent 22 annotations are selected as the labels in our experiments, similar to the settings in~\cite{cao2016deep}. 5000 images are sampled randomly for testing and the rest are regarded as the training and gallery set. 

\textbf{NUS-WIDE Dataset}\footnote{http://lms.comp.nus.edu.sg/research/NUS-WIDE.htm}: This dataset contains 269648 images associated with 81 ground-truth concepts. Hence both the image number and the tag number are much larger than MIRFLICKR-25K and IAPRTC12. There are 222333 images left for the training and testing after the images without any label among the 81 concepts are removed. Following~\cite{zhao2015deep}, we randomly choose 5000 samples as the testing set and the rest are regarded as the gallery and training set.  

\begin{table*}[t]
\centering
\caption{ Comparison of $NDCG@100$ and $ACG@100$ performance among different models on the dataset of MIRFLICKR-25K.}
\begin{tabular}[width =7 in]{|c|c|c|c|c|c|c|c|c|c|c|}
\hline
 \multicolumn{1}{|c|}{} & \multicolumn{5}{c|}{$NDCG@100$} & \multicolumn{5}{c|}{$ACG@100$} \\ \hline
Method & 16 bits & 32 bits & 48 bits & 64 bits & 128 bits & 16 bits & 32 bits & 48 bits & 64 bits & 128 bits \\ \hline
RCDH & 0.4383 & 0.4807 & 0.4827 & 0.4877 & 0.4910 & 2.9189 & 3.0908 & 3.0979 & 3.1145 & 3.1235 \\ \hline
$\textrm{RCDH}_{Norm}$ & 0.4392 & 0.4859 & 0.4926 & 0.4993 & 0.5062 & 2.9196 & 3.1313 & 3.1449 & 3.1801 & 3.2236 \\ \hline
$\textrm{RCDH}_{Interval}$ & 0.4403 & 0.4832 & 0.4811 & 0.4901 & 0.4902 & 2.9284 & 3.1113 & 3.1069 & 3.1375 & 3.1623 \\  \Xhline{1.2pt}
$\textrm{C-RCDH}_{R}$ & 0.4214 & 0.4674 & 0.4768 & 0.4799 & 0.4823 & 2.8749 & 3.0649 & 3.1241 & 3.1655 & 3.1562 \\ \hline
$\textrm{C-RCDH}_{Cla}$ & 0.1408 & 0.1408 & 0.1408 & 0.1408 & 0.1408 & 1.3319 & 1.3319 & 1.3319 & 1.3319 & 1.3319 \\ \hline
$\textrm{C-RCDH}_{Clu}$ & 0.1642 & 0.1642 & 0.1642 & 0.1642 & 0.1642 & 1.1667 & 1.1667 & 1.1667 & 1.1667 & 1.1667 \\ \hline
$\textrm{C-RCDH}_{R+Cla}$ & 0.4412 & 0.4827 & 0.4942 & 0.4987 & 0.5078 & 2.9037 & 3.0979 & 3.1508 & 3.1630 & 3.1962 \\ \hline
$\textrm{C-RCDH}_{R+Clu}$ & 0.4561 & 0.4914 & 0.4978 & 0.5091 & 0.5213 & 2.9568 & 3.1469 & 3.1572 & 3.2355 & 3.2753 \\ \hline
$\textrm{C-RCDH}_{NQ}$ & 0.4360 & 0.4759 & 0.4827 & 0.4860 & 0.5129 & 2.9036 & 3.0573 & 3.1957 & 3.1928 & 3.2393 \\ \Xhline{1.2pt}
C-RCDH & \textbf{0.4786} & \textbf{0.5119} & \textbf{0.5154} & \textbf{0.5284} & \textbf{0.5406} & \textbf{3.1273} & \textbf{3.2499} & \textbf{3.2454} & \textbf{3.3197} & \textbf{3.3679} \\ \hline
\end{tabular}
\label{table:result_mir_a}
\end{table*}

\begin{table*}[t]
\centering
\caption{Comparison of $NDCG@100$ and $ACG@100$ performance among different models on the dataset of IAPRTC12.}
\begin{tabular}[width =7 in]{|c|c|c|c|c|c|c|c|c|c|c|}
\hline
 \multicolumn{1}{|c|}{} & \multicolumn{5}{c|}{$NDCG@100$} & \multicolumn{5}{c|}{$ACG@100$} \\ \hline
Method & 16 bits & 32 bits & 48 bits & 64 bits & 128 bits & 16 bits & 32 bits & 48 bits & 64 bits & 128 bits \\ \hline
RCDH & 0.4212 & 0.4533 & 0.4631 & 0.4707 & 0.4747 & 1.1570 & 1.2257 & 1.2538 & 1.2709 & 1.2733 \\ \hline
$\textrm{RCDH}_{Norm}$ & 0.4247 & 0.4579 & 0.4713 & 0.4835 & 0.4910 & 1.1616 & 1.2264 & 1.2784 & 1.2999 & 1.3018 \\ \hline
$\textrm{RCDH}_{Interval}$ & 0.4217 & 0.4479 & 0.4611 & 0.4752 & 0.4729 & 1.1592 & 1.1862 & 1.2395 & 1.2796 & 1.2604 \\  \Xhline{1.2pt}
$\textrm{C-RCDH}_{R}$ & 0.4149 & 0.4427 & 0.4610 & 0.4713 & 0.4825 & 1.1309 & 1.1840 & 1.2485 & 1.2639 & 1.2700  \\ \hline
$\textrm{C-RCDH}_{Cla}$ & 0.1114 & 0.1114 & 0.1114 & 0.1114 & 0.1114 & 0.3849 & 0.3849 & 0.3849 & 0.3849 & 0.3849 \\ \hline
$\textrm{C-RCDH}_{Clu}$ & 0.1114 & 0.1114 & 0.1114 & 0.1114 & 0.1114 & 0.3850 & 0.3850 & 0.3850 & 0.3850 & 0.3850 \\ \hline
$\textrm{C-RCDH}_{R+Cla}$ & 0.4235 & 0.4560 & 0.4765 & 0.4827 & 0.5001 & 1.1639 & 1.2293 & 1.2616 & 1.2840 & 1.3315 \\ \hline
$\textrm{C-RCDH}_{R+Clu}$ & 0.4296 & 0.4611 & 0.4802 & 0.4973 & 0.5124 & 1.1704 & 1.2417 & 1.2788 & 1.3096 & 1.3554 \\ \hline
$\textrm{C-RCDH}_{NQ}$ & 0.4027 & 0.4422 & 0.4677 & 0.4723 & 0.4910 & 1.1160 & 1.2109 & 1.2570 & 1.2699 & 1.2935 \\ \Xhline{1.2pt}
C-RCDH & \textbf{0.4388} & \textbf{0.4811} & \textbf{0.5050} & \textbf{0.5095} & \textbf{0.5369} & \textbf{1.1897} & \textbf{1.2804} & \textbf{1.3273} & \textbf{1.3477} & \textbf{1.4009}\\ \hline
\end{tabular}
\label{table:result_iap_a}
\end{table*}

\textbf{Evaluation Metrics}: Two popular evaluation metrics for multi-label image retrieval, Normalized Discounted Cumulative Gain (NDCG)~\cite{jarvelin2000ir} and Average Cumulative Gain (ACG)~\cite{jarvelin2000ir}, are used to evaluate the effectiveness of the hashing methods. The metrics can calculate the alignment degree between the order obtained by ground truth and that ranked by hamming distances. 
The formulation of NDCG is: 
\begin{eqnarray}
NDCG@p=\frac{1}{Z}\sum_{i=1}^{p}\frac{2^{r_i}-1}{\mathrm{log}(1+i)},
\end{eqnarray}
in which $Z$ is the normalization factor to keep the value of NDCG equal to $1$ when the ranking order is exactly accurate. In the rank list, the truncated position is denoted as $p$. $r_i$  represents the similarity level of the image on the $i$th position of a ranking list. Specifically, we can formulated the similarity by the annotated labels: $r_i=\mathbf{y}_q^T\mathbf{y}_i$. 
ACG is defined as below:
\begin{eqnarray}
ACG@p=\frac{1}{p}\sum_{i=1}^{p}r_i,
\end{eqnarray}
in which the definitions of $p$ and $r_i$ are the same as NDCG. Two metrics evaluate the effectiveness of hashing methods from different aspects. Compared to NDCG, ACG tends to reveal the similarity degree of top-$p$ images on average. The values of two metrics are both calculated using the hamming distances which are computed by the learned hashing codes.

\begin{figure*}[htbp]
\centering
\subfigure[]{
\begin{minipage}[t]{0.33\linewidth}
\centering
\includegraphics[width=6cm]{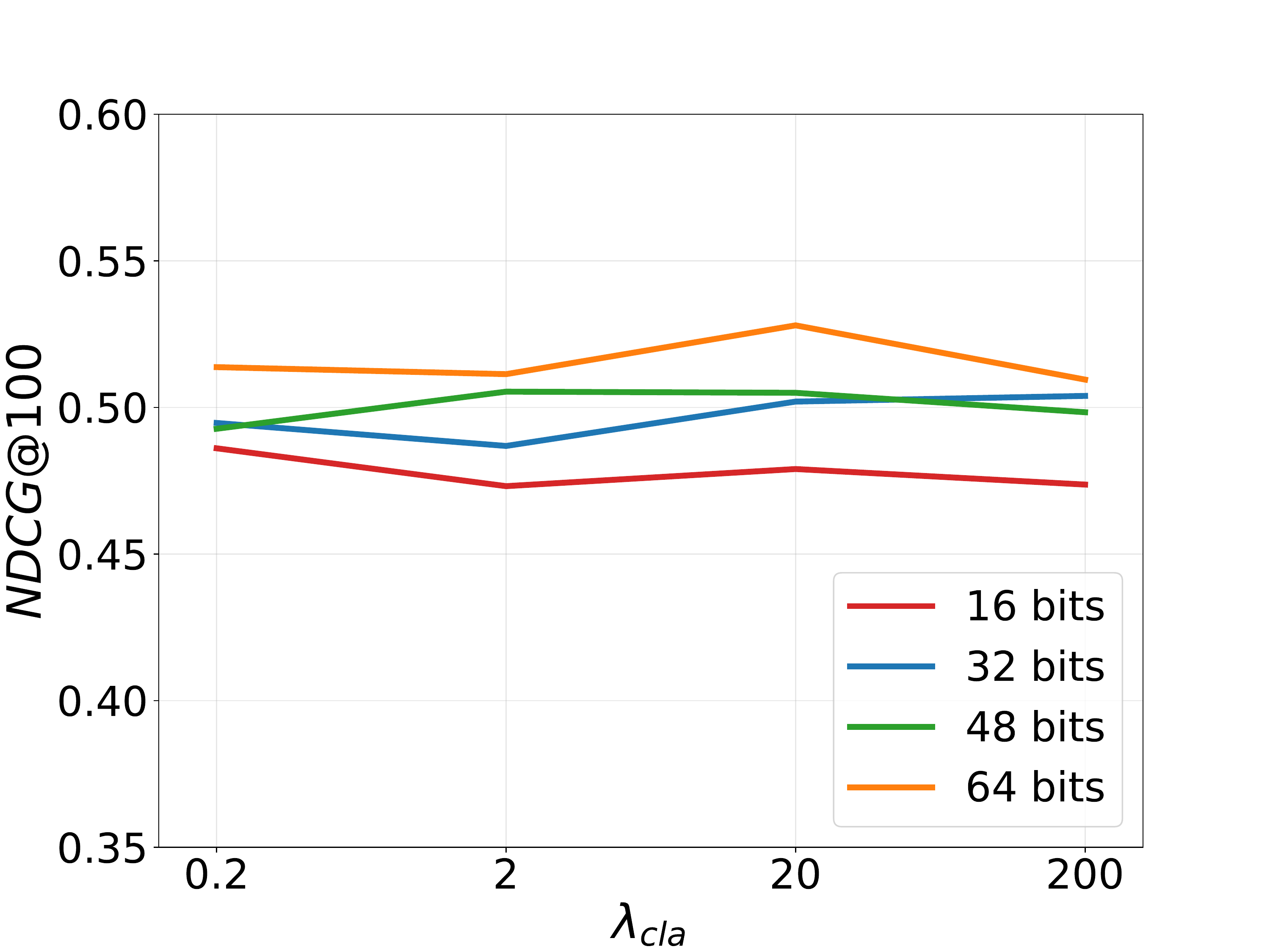} \\
\vspace{-3mm}
\includegraphics[width=6cm]{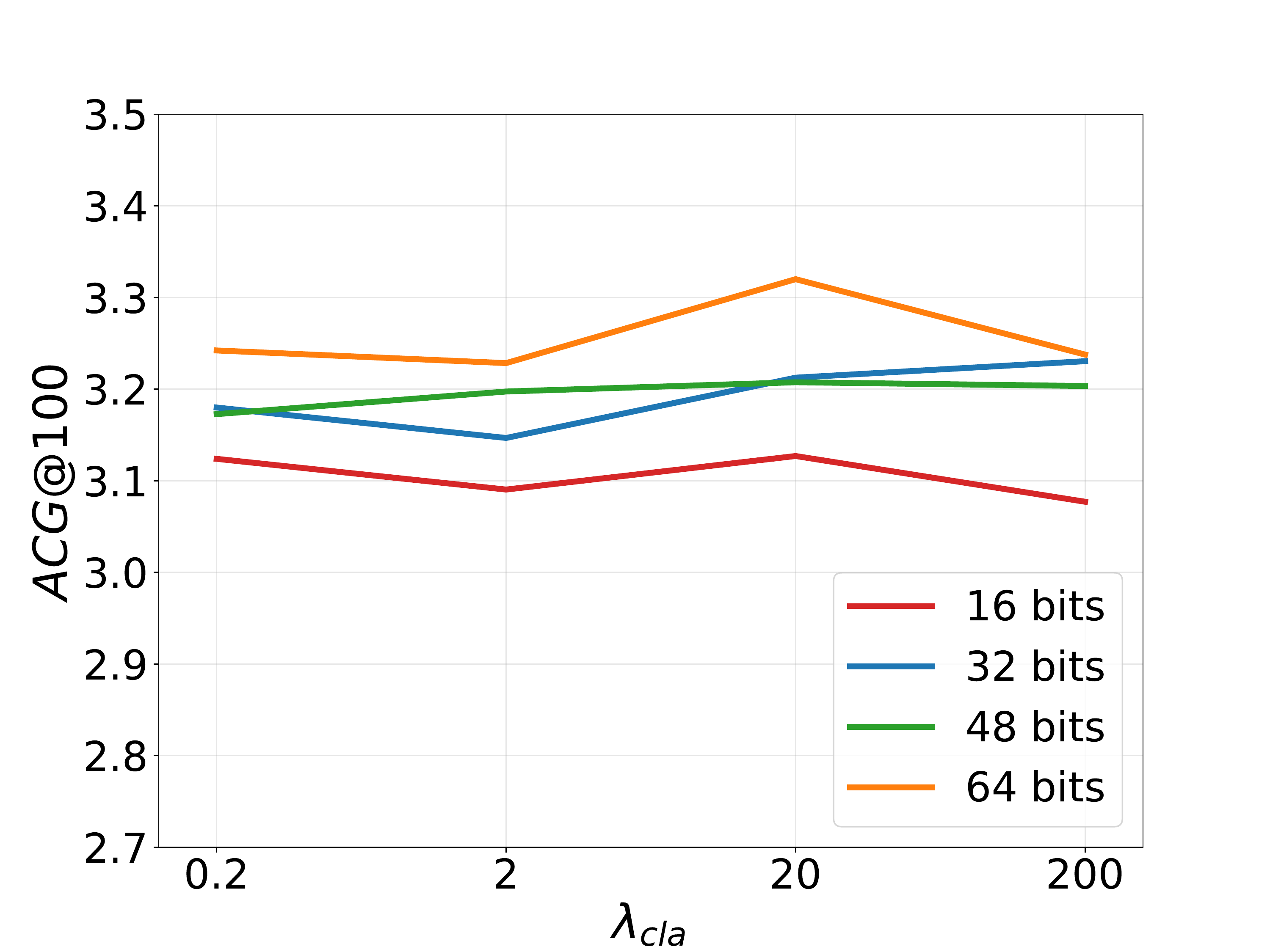}
\end{minipage}%
}%
\subfigure[]{
\begin{minipage}[t]{0.33\linewidth}
\centering
\includegraphics[width=6cm]{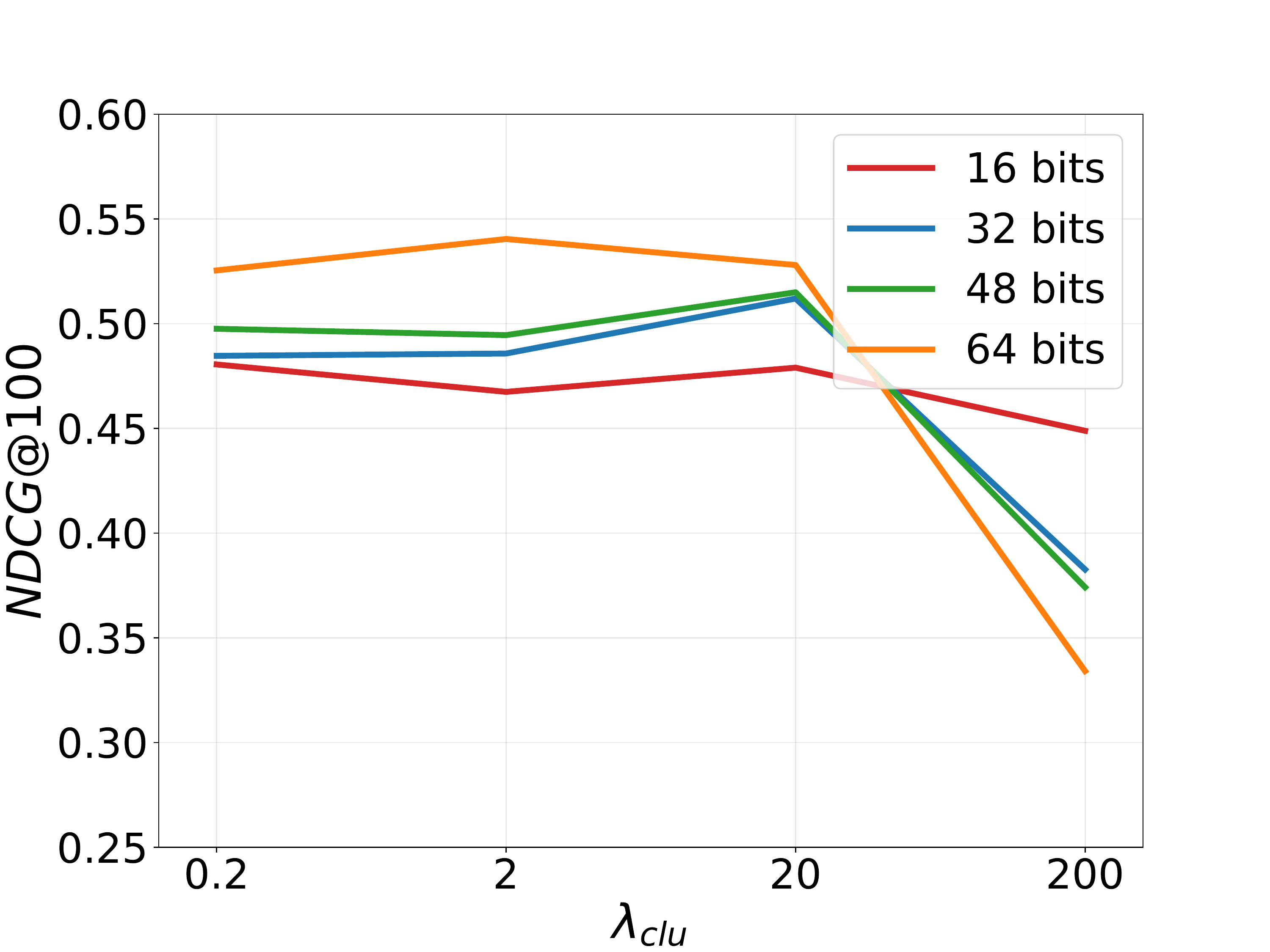} \\
\vspace{-3mm}
\includegraphics[width=6cm]{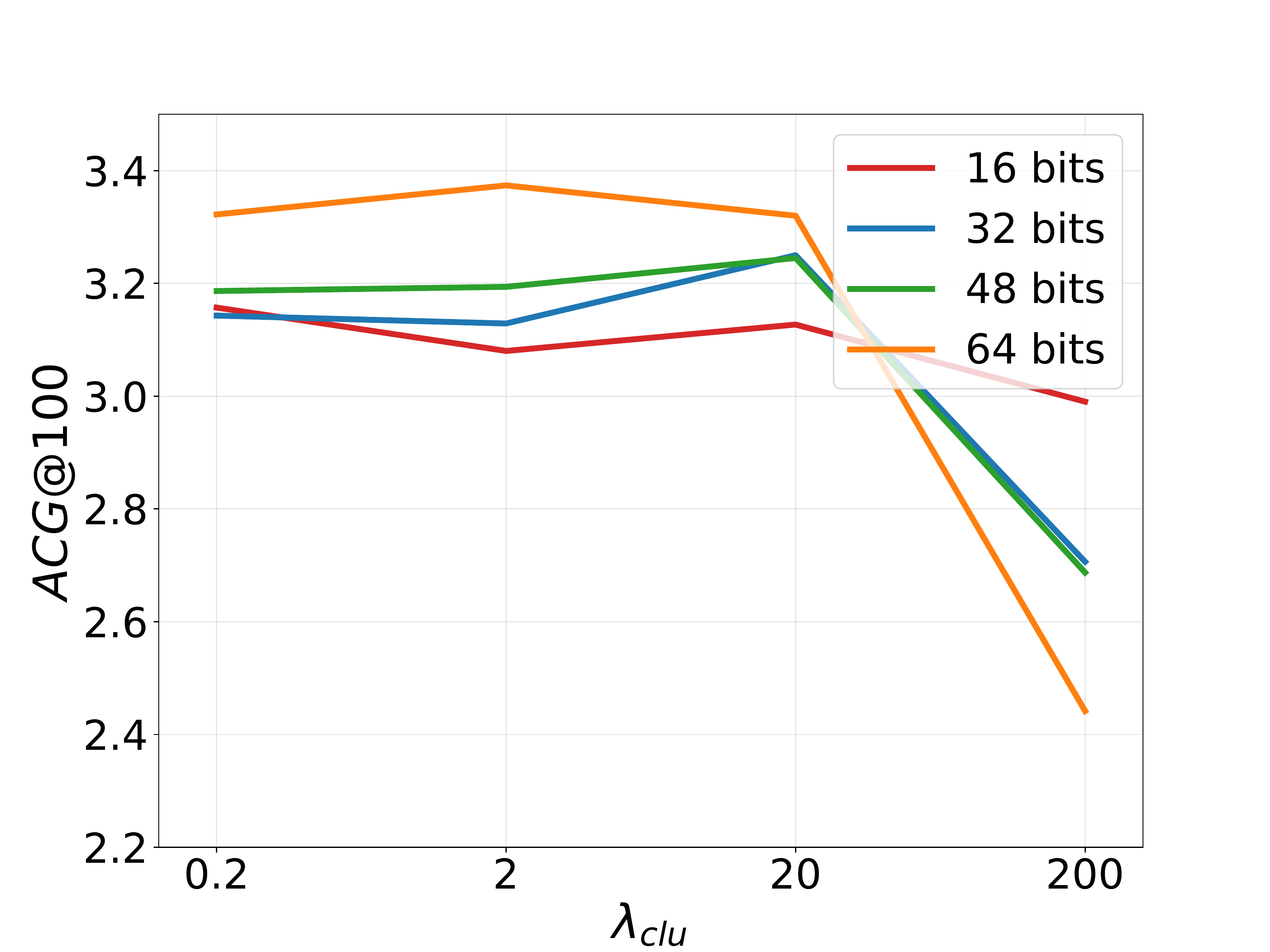}
\end{minipage}%
}%
\subfigure[]{
\begin{minipage}[t]{0.33\linewidth}
\centering
\includegraphics[width=6cm]{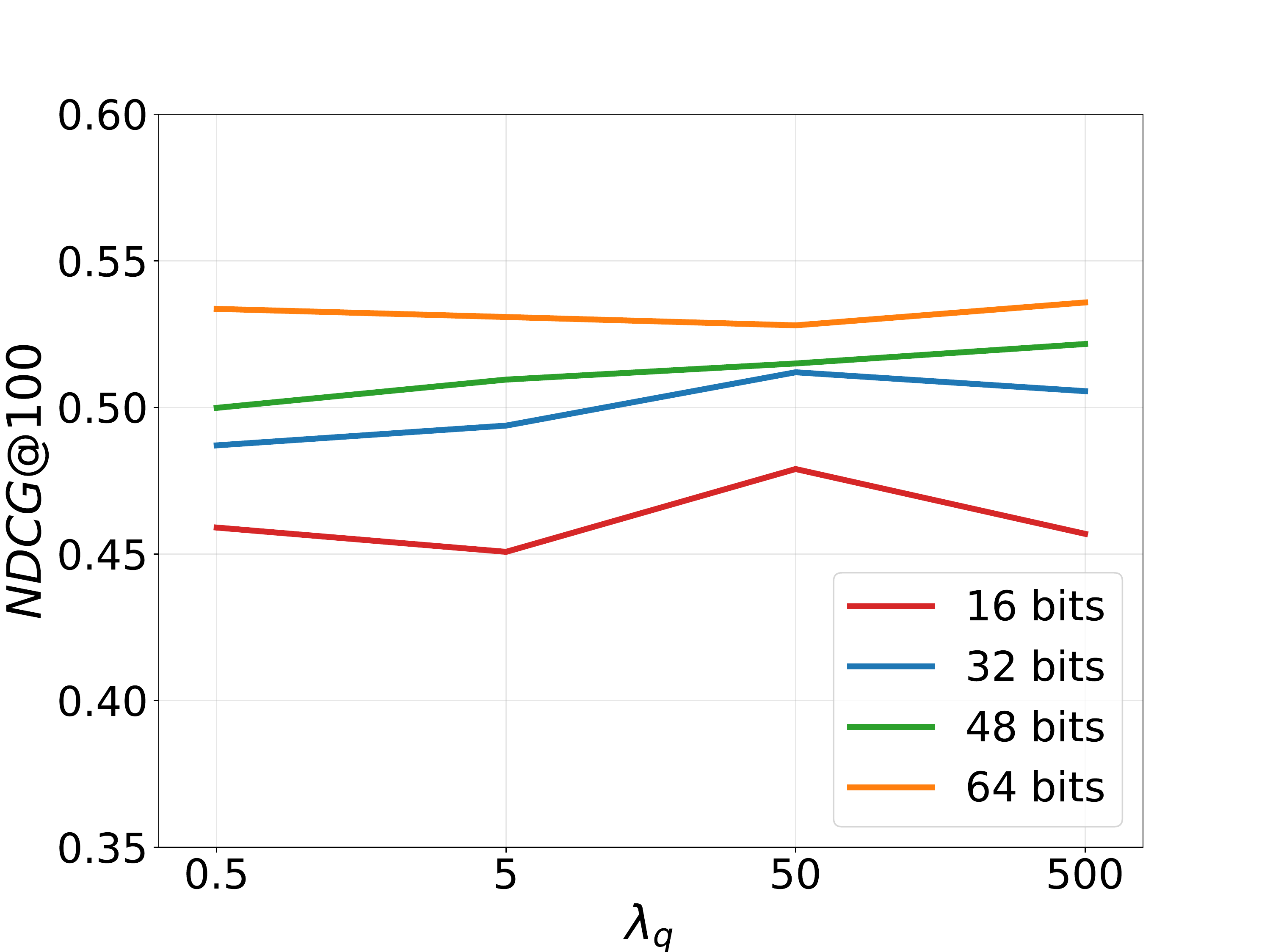} \\
\vspace{-3mm}
\includegraphics[width=6cm]{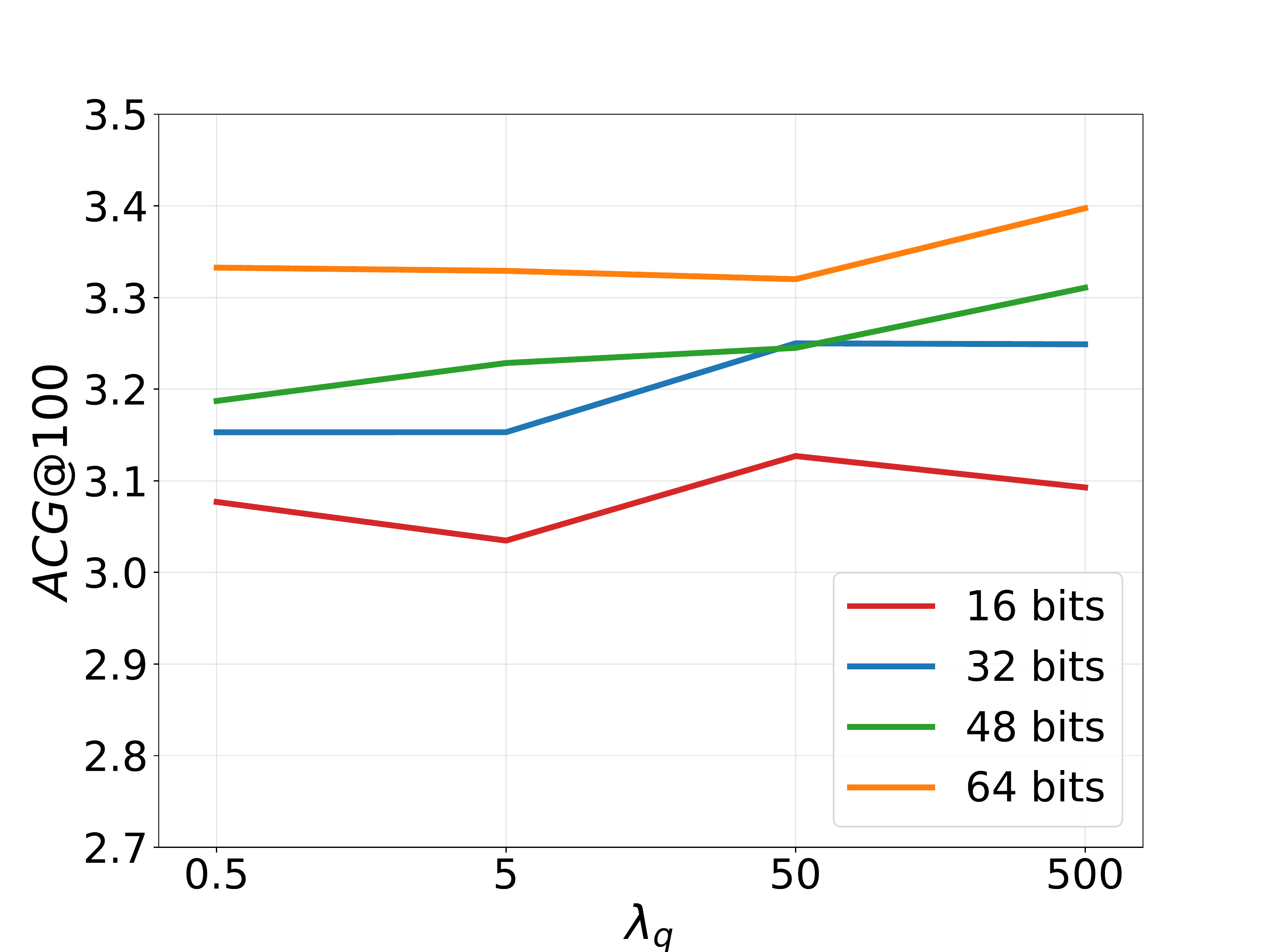}
\end{minipage}
}
\centering
\caption{Experimental results of parameter sensitivity on MIRFLICKR-25K. The first row presents the $NDCG@100$ values while the $ACG@100$ values are in the second row. The results of $\lambda_{cla}$, $\lambda_{clu}$ and $\lambda_q$ are shown in (a), (b) and (c), respectively. }
\label{fig:param_sen}
\vspace{-3mm}
\end{figure*}

\subsection{Results and Analysis}

\textbf{Comparison with Conventional Hashing Methods}: We compared our C-RCDH to seven conventional hashing methods including PCA Hashing (PCA-H)~\cite{wang2010semi} PCA Iterative Quantization (PCA-ITQ)~\cite{gong2013iterative}, Sequential Projection Learning Hashing (SPLH)~\cite{wang2010semi}, Kernel Supervised Hashing (KSH)~\cite{liu2012supervised}, Column-Generation Hashing (CGH)~\cite{li2013learning}, Fast Hashing (FH)~\cite{lin2015supervised}, Supervised Discrete Hashing (SDH)~\cite{shen2015supervised} and CCA Iterative Quantization (CCA-ITQ)~\cite{gong2013iterative}. In the experiments, we used the output representation from the layer fc7 in the pretrained VGG neural networks for a fair comparison. PCA-ITQ is unsupervised and CCA-ITQ is the supervised extension of it. KSH employs the kernel formulation to produce discriminative hashing codes. In our experiments, we chose RBF Gaussian kernel for KSH. In order to maintain the relation contained in images, CGH exploits the advantages of column generation and triplet training. For each data point, we chose 50 positive samples and 50 negative samples to build the triplet set. FH utilizes boosted decision trees to learn hashing codes on a large-scale training set, and we inferred the binary codes by hinge loss function. Meanwhile, SDH alleviates the regularization issue using cyclic coordinate descent. In our implementation, the kernel version was selected and $l_2$-loss was used for optimization.

The performance of $NDCG@100$ and $ACG@100$ on the datasets of MIRFLICKR-25K, IAPRTC12 and NUS-WIDE is presented in Table~\ref{table:result_mir_s}, Table~\ref{table:result_iap_s} and Table~\ref{table:result_nus_s}. It is clear that for each length of hashing codes, our method obtains much better performance than other methods. The methods with kernel model or linear projection may not handle the distribution of multi-label images exactly due to the inadequate representation ability. The methods of SPLH and FH utilize the pairwise similarity as supervision signals, which cannot achieve comparable performance with our method. The triplets built in the optimization process of CGH are also inadequate to provide holistic guidance although CGH exploits the rank-based supervision information. CCA-ITQ has the best retrieval results among the compared methods. The reason is that the similarity is preserved in binary codes by canonical correlation analysis (CCA). However, the combination of the rank-consistency loss and multi-label classification and clustering loss improves the performance by about 30$\%$-45$\%$ on MIRFLICKR-25K, 60$\%$-70$\%$ on IAPRTC12 and 20$\%$-35$\%$ on NUS-WIDE. Hence the improvement shows the effectiveness of our method. 

\begin{figure*}[t]
  \centering
  \centerline{\epsfig{figure=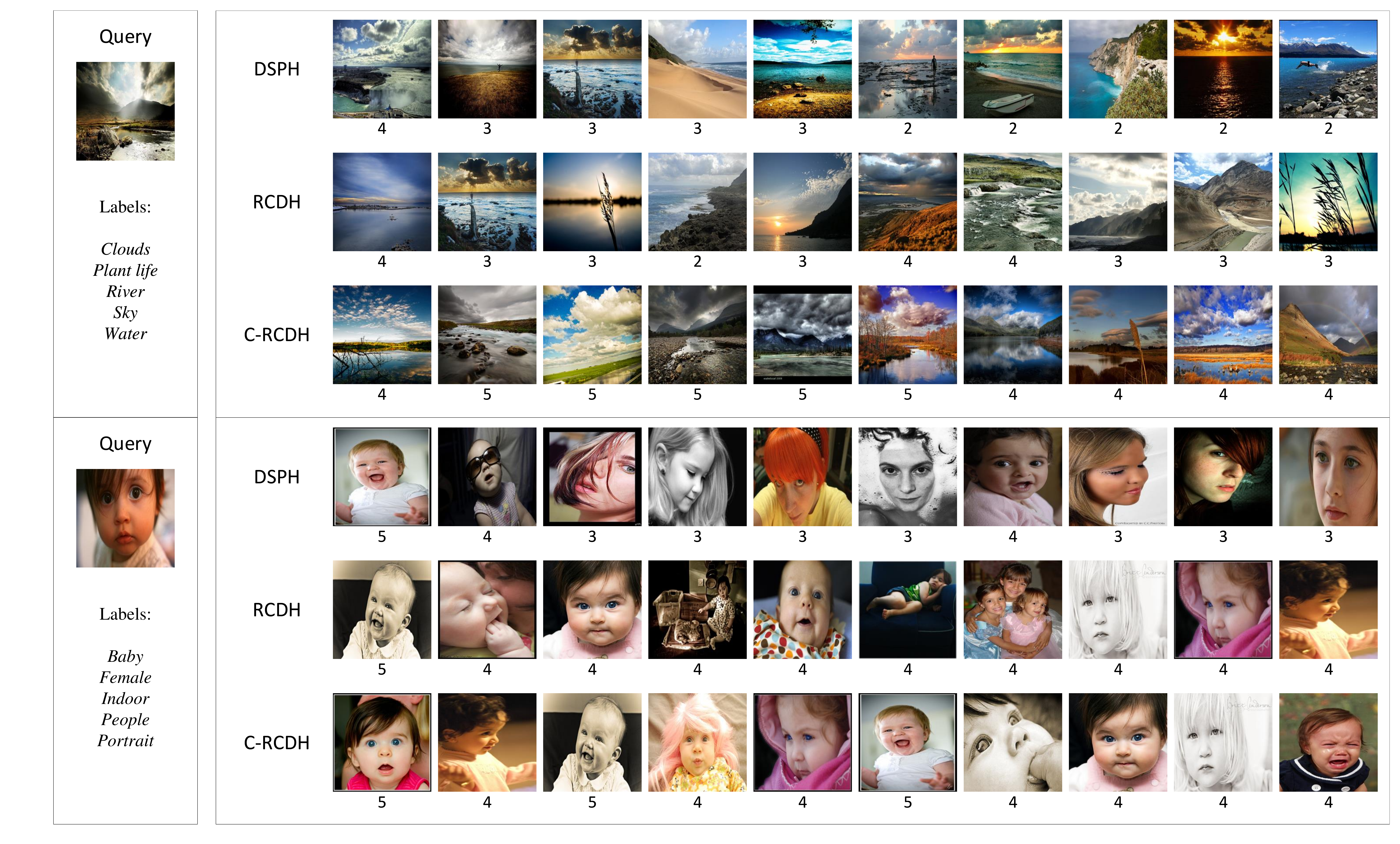,width=18cm}}
  \vspace{-3mm}
\caption{Retrieval comparison with multi-label deep hashing methods on the MIRFLICKR-25K dataset. Top 10 retrieved images by 32-bit hashing codes are displayed for each method. The number below each retrieved image represents its common label number with the query image. The retrieved images of our C-RCDH method are more similar to the query images than those of other methods. }
\label{fig:visualization}
\vspace{-3mm}
\end{figure*}

\textbf{Comparison with Deep Multi-Label Hashing Methods}: Moreover, we compared C-RCDH with other deep multi-label hashing methods, for example, deep supervised hashing with pairwise labels (DPSH)~\cite{li2015feature}, deep supervised hashing with triplet labels (DTSH)~\cite{wang2016deep}, deep semantic ranking based hashing (DSRH)~\cite{zhao2015deep}, instance similarity deep hashing (ISDH)~\cite{zhang2018instance}, deep semantic-preserving hashing (DSPH)~\cite{yao2016deep}, multi-label deep sparse hashing (MDSH)~\cite{liong2018multi} and rank-consistency multi-label deep hashing (RCDH)~\cite{8486592}. Results of $NDCG@100$ and $ACG@100$ on the datasets of MIRFLICKR-25K, IAPRTC12 and NUS-WIDE are presented in Table~\ref{table:result_mir_d}, Table~\ref{table:result_iap_d} and Table~\ref{table:result_nus_d} respectively. It is indicated that C-RCDH achieves state-of-the-art retrieval performance. DPSH and DTSH maximize the likelihood of sampled pairwise labels or triplet labels and extract hashing codes by deep neural networks. 
The reason that DPSH and DTSH do not yield competitive performance is that their abilities to exploit complex semantic structure is not powerful enough. The different levels of similarity for multi-label images may not be captured by the objectives. 
In ISDH, while the pairwise similarity is redefined by normalizing semantic labels, the proposed pairwise loss is limited in providing powerful guidance to generate effective hashing codes.
The results of RCDH on short binary codes are better than DSRH, which infers that the rank-consistency loss is more efficient in learning short binary codes. The convergence of DSRH and DSPH needs considerable triplets, resulting that it need longer hashing vectors to achieve excellent performance. 
Additionally, the triplet supervision may lead to local optimum since that it may not provide global enough training signals to the deep neural networks. However, the RCDH and C-RCDH methods supervise the learning process by effective ranking lists on every code length, so that we can also get pleasant retrieval results using short vectors. Furthermore, the proposed C-RCDH has better retrieval performance with long hashing codes than RCDH since that we have designed a new loss function to penalize the samples which have inconsistent similarity in the original space and the hamming space. The normalization technique guarantees enough training signals for the back-propagation process. Besides, the formulation is easy to implement while the retrieval results are satisfactory. The C-RCDH model jointly trained by the above mentioned objective functions has improved the performance over RCDH by about 6$\%$-9$\%$ on MIRFLICKR-25K, 4$\%$-9$\%$ on IAPRTC12 and 6$\%$-10$\%$ on NUS-WIDE, especially 3$\%$ more improvment on large numbers of hashing bits. Hence the comparison with state-of-the-art multi-label hashing methods demonstrates the efficiency and effectiveness of the proposed method.

\textbf{Ablation Study}: To evaluate the effects of each part in the proposed method, we obtained different models by using different objective functions for the training process. To evaluate the modification of the rank-consistency loss, we compared the models with just one modification with the original model of RCDH. The model utilizing the new method to determinate the penalty intervals is denoted as $\textrm{RCDH}_{Interval}$, while the model with the normalization formulation is called $\textrm{RCDH}_{Norm}$. Note that the model of RCDH also makes use of the multi-label classification loss. Hence the model with complete modifications on the rank-consistency loss is equivalent to the C-RCDH model whose multi-label clustering loss is removed. We denote such model as $\textrm{C-RCDH}_{R+Cla}$. Similarly, the other compared models are defined as follows: $\textrm{C-RCDH}_{R}$ is the model whose multi-label classification and clustering losses are both removed; $\textrm{C-RCDH}_{Cla}$ is the model with only the classification and quantization losses; $\textrm{C-RCDH}_{Clu}$ is the model with only the clustering and quantization losses; $\textrm{C-RCDH}_{R+Clu}$ is the model without the classification loss. The quantization loss is used in the training process of all the above models. $\textrm{C-RCDH}_{NQ}$ is the model without the quantization loss. Hence from the above designed experiments, we are able to infer the importance of each term in the final objective function. 

The experimental results on two datasets, MIRFLICKR-25K and IAPRTC12 are shown in Table~\ref{table:result_mir_a} and Table~\ref{table:result_iap_a}. From the tables we can see that when the normalization technique is added to the rank-consistency loss, there is little difference between $\textrm{RCDH}_{Norm}$ with RCDH when hashing codes are short. As $\gamma$ is set to 16, $\textrm{RCDH}_{Norm}$ is just the same as the model of RCDH without the normalization technique on 16 bits. However, when the hashing codes become longer, the improvement of $\textrm{RCDH}_{Norm}$ is more and more significant. The back-propagated signals will not disappear with the new penalty loss function. Hence we develop the merits of the original model on short bits and obtain encouraging performance on all the lengths of hashing codes. Moreover, we can find that $\textrm{RCDH}_{Interval}$ has a slight improvement on RCDH. This indicates that the partition of hamming space also influences the learning of binary codes, but not significantly. However, the upper and lower bounds of the intervals are easier to determine, which promotes the efficiency of the implementation. Therefore, it is reasonable that the model with the both modifications, i.e., $\textrm{C-RCDH}_{R+Cla}$ can outperform both $\textrm{RCDH}_{Norm}$ and $\textrm{RCDH}_{Interval}$. 

From Table~\ref{table:result_mir_a} and Table~\ref{table:result_iap_a}, it can also be observed that the models of $\textrm{C-RCDH}_{Cla}$ and $\textrm{C-RCDH}_{Clu}$ both lose the ability to search similar images. The reason can be inferred from the illustration of Fig.~\ref{fig:visual}. When the model is trained by only the multi-label classification loss, the distribution of features may be separable for different categories. However, the distribution is not suitable for the retrieval using binarized codes. Besides, if the model is trained by only the multi-label clustering loss, it is expected that all the related samples will gather to one point according to the analyses mentioned above. Hence $\textrm{C-RCDH}_{Clu}$ also fails to give valid retrieval results. As for the models with the rank-consistency loss, $\textrm{C-RCDH}_{R}$ presents pleasing results although there is no classification and clustering losses to manipulate the distribution of learned features. However, $\textrm{C-RCDH}_{R+Cla}$ obtains better performance than $\textrm{C-RCDH}_{R}$, which demonstrates the necessity of the multi-label classification loss. Besides, it is encouraging that $\textrm{C-RCDH}_{R+Clu}$ achieves better performance than $\textrm{C-RCDH}_{R+Cla}$ for the reason that the clustering loss can help to manipulate the neighborhood structure of learned features associated with different labels. This term can work even though the classification loss is removed, which proves the effectiveness of the proposed method. From the results given by $\textrm{C-RCDH}_{NQ}$, we can see the removal of the quantization loss
has an impact on the learning of hashing codes. The calculation of hamming distances in our formulation relies on the assumption that the real-valued outputs of the deep neural networks should be close to their corresponding binary codes. As a result, this term is also indispensable in the training process. In conclusion, each term in the final objective function of C-RCDH plays an important role and an appropriate combination yields the compelling retrieval performance of C-RCDH.

\textbf{Parameter Sensitivity}: We implement the experiments on the dataset of MIRFLICKR-25K with hashing codes of 16, 32, 48 and 64 bits. We tune the parameters of $\lambda_{cla}$, $\lambda_{clu}$ and $\lambda_q$ by a factor of 10. The $NDCG@100$ and $ACG@100$ comparisons are shown in Fig.~\ref{fig:param_sen}. From the figures, we see the retrieval performance is not sensitive to $\lambda_{cla}$ and $\lambda_q$. While the two terms of losses, the multi-label classification loss and the quantization loss, are important to the final optimum, a small or large value of $\lambda_{cla}$ and $\lambda_q$ will not severely affect the performance. However, for $\lambda_{clu}$, we see the values of $NDCG@100$ and $ACG@100$ drop a lot when $\lambda_{clu}$ achieves 200, which means that a too large value of $\lambda_{clu}$ can indeed influence the training process. As analyzed in Section~\ref{sec:necessity}, the reason is that when the weight of the multi-label clustering loss is too large, all the related vectors and center vectors will get closer and closer. In this case, the discriminative ability of the vectors may be reduced. Hence the comparison results of parameter sensitivity also verifies our analysis experimentally. 

\textbf{Visualization}: We extracted 32-bit hashing codes by three multi-label deep hashing methods, DSPH, RCDH and C-RCDH, to retrieve similar images on MIRFLICKR-25K. The examples of top 10 returned results are shown in Fig.~\ref{fig:visualization}. The number below each retrieved image represents its common label number with the query image. From the figure, we see that the retrieved images of our C-RCDH methods are more similar to the query images than those of other methods. 
For the first query image, our method can retrieve images with the label \emph{River} while other methods retrieve images which only have the label \emph{Water} and maybe \emph{Sea} or \emph{Lake} without containing \emph{River}. For the second query image, our C-RCDH method can also successfully capture all the five semantics and give pleasing searching results. Therefore our method can generate effective hashing codes to accomplish accurate multi-label image retrieval where not only the foreground objects of the returned images are corresponding to the query, but the background features are also exactly captured. 
Hence the visualization results validate the effectiveness of our proposed method. 

\section{Conclusion}

In this paper, we have proposed a new rank-consistency deep hashing method for scalable multi-label image search. In order to produce effective hashing codes, we have divided the hamming space to multiple intervals according to the categories partitioned by the original similarity degrees. Then we have supervised the training by listwise alignment so as to keep the rank-consistency between two spaces. Moreover, to manipulate the distribution of the features, we have proposed a multi-label softmax cross-entropy loss and a multi-label clustering loss. The two terms have been added to the objective function to train the model jointly for the purpose of strengthening the discriminative power of the learned features. With such supervision, the features have gained a polymerized neighborhood structure to achieve the goal that the distances in hamming space should be corresponding to the semantic similarity in the original space. Hence different terms in the objective function influence each other and the deep network can be optimized efficiently. Retrieval results on three public multi-label datasets have demonstrated the superior performance of our proposed approach.



{\small
\bibliographystyle{IEEEtran}
\bibliography{bare_jrnl}
}
%

\newcommand{\controlspace}{\vspace{-10pt}}
\controlspace

\begin{IEEEbiography}[{\includegraphics[width=1in,height=1.25in,clip,keepaspectratio]{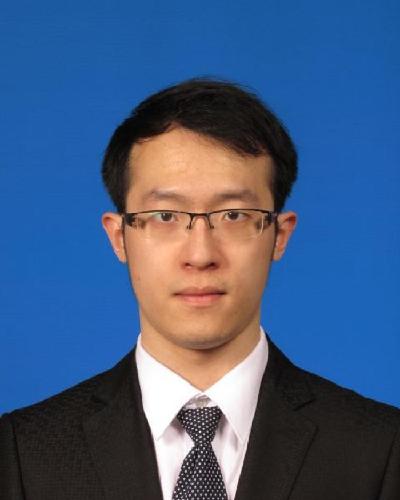}}]{Cheng Ma}
received the B.Eng. degree from the Department of Electronic Engineering, Tsinghua University, China, in 2017. He is currently working toward the Ph.D. degree in the Department of Automation, Tsinghua University, China. His research interests include image retrieval, super resolution and generative adversarial networks.
\end{IEEEbiography}

\vfill
\controlspace

\begin{IEEEbiography}[{\includegraphics[width=1in,height=1.25in,clip,keepaspectratio]{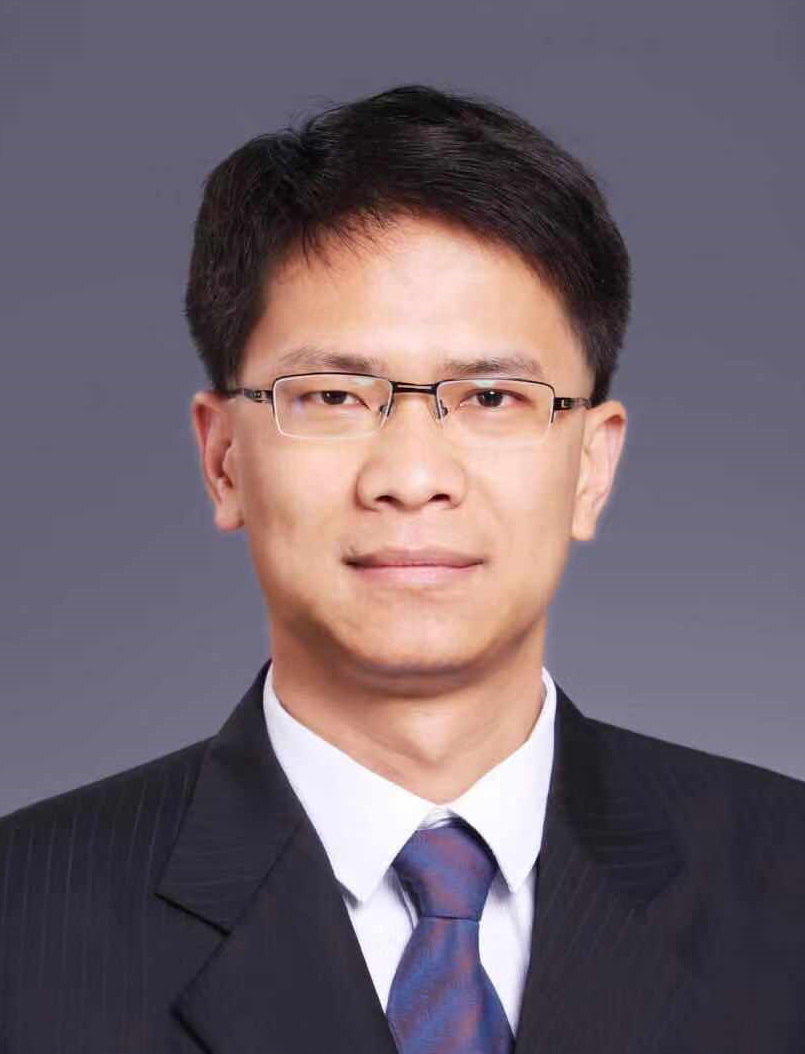}}]{Jiwen Lu}
(M’11-SM’15) received the B.Eng. degree in mechanical engineering and the M.Eng. degree in electrical engineering from the Xi’an University of Technology, Xi’an, China, in 2003 and 2006, respectively, and the Ph.D. degree in electrical engineering from Nanyang Technological University, Singapore, in 2012. He is currently an Associate Professor with the Department of Automation, Tsinghua University, Beijing, China. His current research interests include computer vision, pattern recognition, and machine learning. He has authored/co-authored over 250 scientific papers in these areas, where 70+ of them are the IEEE Transactions papers and 60 of them are CVPR/ICCV/ECCV/NeurIPS papers. He serves the Co-Editor-of-Chief of the Pattern Recognition Letters, an Associate Editor of the IEEE Transactions on Image Processing, the IEEE Transactions on Circuits and Systems for Video Technology, the IEEE Transactions on Biometrics, Behavior, and Identity Science, and Pattern Recognition. He was/is a member of the Multimedia Signal Processing Technical Committee and the Information Forensics and Security Technical Committee of the IEEE Signal Processing Society, and a member of the Multimedia Systems and Applications Technical Committee and the Visual Signal Processing and Communications Technical Committee of the IEEE Circuits and Systems Society. He is a senior member of the IEEE.
\end{IEEEbiography}

\vfill
\controlspace

\begin{IEEEbiography}[{\includegraphics[width=1in,height=1.25in,clip,keepaspectratio]{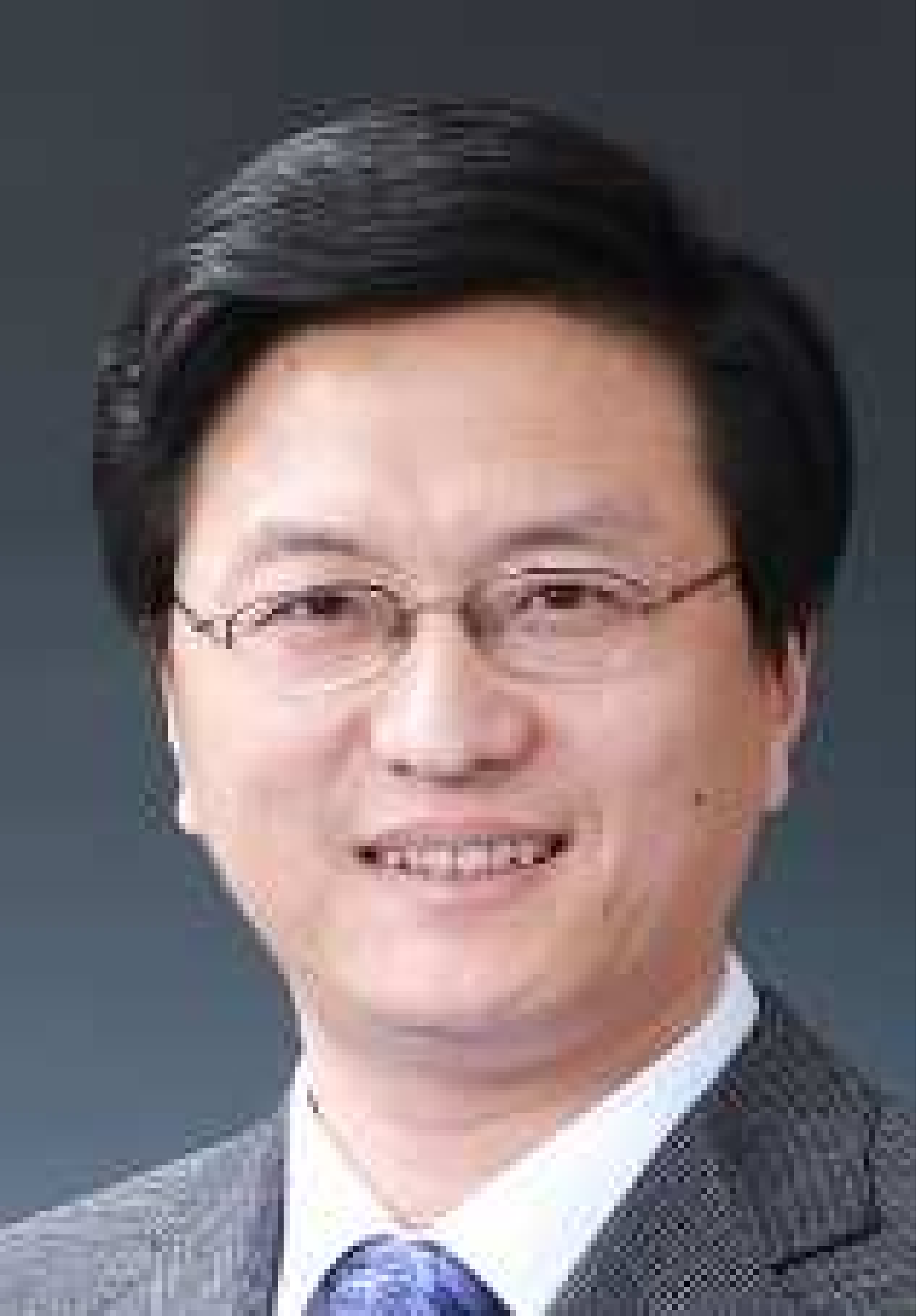}}]{Jie Zhou}
(M’01-SM’04) received the BS and MS degrees both from the Department of Mathematics, Nankai University, Tianjin, China, in 1990 and 1992, respectively, and the PhD degree from the Institute of Pattern Recognition and Artificial Intelligence, Huazhong University of Science and Technology (HUST), Wuhan, China, in 1995. From then to 1997, he served as a postdoctoral fellow in the Department of Automation, Tsinghua University, Beijing, China. Since 2003, he has been a full professor in the Department of Automation, Tsinghua University. His research interests include computer vision, pattern recognition, and image processing. In recent years, he has authored more than 100 papers in peer-reviewed journals and conferences. Among them, more than 30 papers have been published in top journals and conferences such as the IEEE Transactions on Pattern Analysis and Machine Intelligence, IEEE Transactions on Image Processing, and CVPR. He is an associate editor for the IEEE Transactions on Pattern Analysis and Machine Intelligence and two other journals. He received the National Outstanding Youth Foundation of China Award. He is a senior member of the IEEE.\end{IEEEbiography}

\vfill

\end{document}